\documentclass[twoside,11pt]{article}

\usepackage{graphicx} 
\usepackage{amsmath,amsfonts,amssymb}
\usepackage{url,algorithm}
\usepackage{color,subfigure}
\usepackage{multirow}
\usepackage{rotating}
\usepackage{natbib}
\usepackage{subfigure}

\newcommand{\qed}{\hfill$\Box$\vspace{0em}\par\noindent}

\renewcommand{\eqref}[1]{(\ref{#1})}

\graphicspath{{figures/}}

\ifx\thedefinition\undefined
\newtheorem{definition}{Definition}
\fi
\ifx\thelemma\undefined
\newtheorem{lemma}{Lemma}
\fi
\ifx\thetheorem\undefined
\newtheorem{theorem}{Theorem}
\fi
\ifx\proof\undefined
\newcommand{\proof}{\noindent{\em Proof}.~}
\fi

\usepackage{enumitem}

\newcommand{\rr}{{\mathbb R}}
\newcommand{\ba}[1]{\begin{array}{#1}}
\newcommand{\ea}{\end{array}}
\newcommand{\beqar}{\begin{eqnarray}}
\newcommand{\eeqar}{\end{eqnarray}}
\newcommand{\st}{\mathop{\rm s.t.}\nolimits}
\newcommand{\card}{\mathop{\rm card}\nolimits}

\newcommand{\XX}{{\mathcal X}}
\newcommand{\YY}{{\mathcal Y}}
\newcommand{\PP}{{\mathcal P}}
\newcommand{\CC}{{\mathcal C}}
\newcommand{\BB}{{\mathcal B}}

\ifx\thelemma\undefined
\newtheorem{lemma}{Lemma}
\fi
\ifx\thecorollary\undefined

\fi
\newcommand{\smallmat}[1]{\left[ \begin{smallmatrix}#1 \end{smallmatrix} \right]}

\begin{document}

\title{Piecewise linear regression and classification}

\author{Alberto Bemporad$^\dagger$}
\date{{\small
        $^\dagger$IMT School for Advanced Studies Lucca, Italy \\Email: 
        \texttt{alberto.bemporad@imtlucca.it}}\\[3em]%
    \today
}

\maketitle

\begin{abstract}
This paper proposes a method for solving multivariate regression and 
classification problems using
piecewise linear predictors over a polyhedral partition of the feature space. 
The resulting
algorithm that we call PARC (Piecewise Affine Regression and Classification)
alternates
between ($i$) solving ridge regression problems for numeric targets, softmax 
regression problems for categorical targets, and either softmax regression or 
cluster centroid computation for piecewise 
linear separation, and ($ii$) assigning the training points to different 
clusters on the basis
of a criterion that balances prediction accuracy and piecewise-linear 
separability.
We prove that PARC is a block-coordinate descent algorithm that optimizes a 
suitably constructed
objective function, and that it converges in a finite number of steps to a 
local 
minimum of that function.
The accuracy of the algorithm is extensively tested numerically on synthetic 
and real-world datasets,
showing that the approach provides an extension of linear 
regression/classification that is
particularly useful when the obtained predictor is used as part of an 
optimization model. A Python implementation of the algorithm described in this 
paper is available at \url{http://cse.lab.imtlucca.it/~bemporad/parc}.

\textbf{Keywords}: Multivariate regression, multi-category classification,
    piecewise linear functions, softmax regression, mixed-integer 
    programming
    
\end{abstract}

\section{Introduction}

Several methods exist for solving supervised learning problems of regression 
and classification~\citep{HTF09,Bis06}. The main goal is to estimate a model of 
the 
data generation process to \emph{predict} at best the target value 
corresponding to a combination of 
features not seen before. However, not all methods are suitable to 
\emph{optimize} on top of the estimated 
model, i.e., to solve a mathematical programming problem that contains the 
estimated model as part of the 
constraints and/or the objective function. For example, to find the best 
combination of features providing a 
desired target, possibly under constraints on the features one can choose. In 
this case, the model is used as a 
surrogate of the underlying (and unknown) features-to-target mapping to 
formulate the decision problem. 
Applications range from derivative-free black-box 
optimization~\citep{Kus64,Jon01,BCD10,Bem20,BP21},
to engineering design~\citep{QHSGVT05}, and control engineering, in 
particular model predictive 
control~\citep{CB99,MRD18,BBM17}, where actuation commands are decided in 
real-time 
by a 
numerical optimization
algorithm based on a dynamical model of the controlled process that is learned 
from 
data~\citep{Lju99,SL19}, see for instance the approach proposed recently 
in~\citep{MB21}.

When optimizing over a learned model is a goal, a clear tradeoff exists between 
the accuracy
of the model on test data and the complexity of the model, which ultimately 
determines the
complexity of the mathematical programming problem resulting from using the 
model. On one extreme, 
we have 
linear regression models, which are very simple to represent as linear 
relations among 
optimization variables but have limited expressiveness. On the other extreme, 
random forests
and other ensemble methods, k-nearest neighbors, kernel support vector machines,
and other methods, can capture
the underlying model very accurately but are difficult to encode in an 
optimization problem.
Neural networks and Gaussian processes can be a good compromise between the
compactness of the model and the
representation of the feature-to-target relation, but are nonlinear models 
leading to nonconvex optimization problems that are possibly
difficult to solve to global optimality.

In this paper, we advocate the use of \emph{piecewise linear} (PWL) models as a 
good
tradeoff between their simplicity, due to the linearity of the model on 
polyhedral regions
of the feature-vector space, and expressiveness, due to the good approximation 
properties of piecewise linear functions~\citep{Bre93,LU92,CD88,JDD00,BOPS11}. 
We 
refer to such models with the more appropriate, although less common, term
\emph{piecewise affine} (PWA), to highlight the presence of an intercept in 
each submodel. PWA models can be easily encoded into optimization problems by 
using mixed-integer linear inequalities~\citep{BM99}, and hence optimize over 
them to reach a global minimum
by using mixed-integer programming~\citep{Lod10}, for which excellent public 
domain and
commercial packages exist.

Many classical machine learning methods have an underlying PWA structure: ridge 
classification,
logistic (and more generally softmax) regression,  hinging 
hyperplanes~\citep{Bre93}, and neural networks with ReLU 
activation functions, they all require evaluating the maximum of linear 
functions to predict 
target values; the predictor associated with a decision tree is a piecewise 
constant (PWC) function
over a  partition of the feature-vector space in boxes; $k$-nearest neighbor 
classifiers can be 
also expressed as PWC functions over polyhedral partitions (the comparison of 
squared Euclidean norms 
$\|x-x_i\|_2^2\leq \|x-x_j\|_2^2$ used to determine the nearest neighbors of 
$x$ 
is equivalent to the linear relation $2(x_j-x_i)'x\leq 
\|x_j\|_2^2-\|x_i\|_2^2$), although the number of 
polyhedra largely grows with the number of training samples. 

Different piecewise affine regression methods have been proposed in the 
system identification literature for getting switching linear dynamical models 
from data~\citep{FMLM03a,RBL04a,BGPV05,NTK05,HLCXV15}.
See also the survey paper~\citep{PKFV07} and the recursive PWA regression 
algorithms proposed 
in~\citep{BBD11,BPB16a}. Most of such methods
identify a prescribed number of linear models and associate one of them to each 
training 
datapoint,
therefore determining a clustering of the data. As a last step,
a multicategory discrimination problem is solved to determine a function
that piecewise-linearly separates the clusters~\citep{BM94}. For instance, the 
approach
of~\cite{NTK05} consists of first clustering the feature+target vectors by
using a Gaussian mixture model, then use support vector classification
to separate the feature-vector space. In~\citep{FMLM03a}, the authors propose
instead to cluster the vectors whose entries are the 
coefficients of local linear models, one model per datapoint, then 
piecewise-linearly separate the clusters. In~\citep{BPB16a}, $K$ recursive 
least-squares 
problems for regression are run in parallel to cluster data in on-line fashion,
based on both quality of fit obtained by each linear model and proximity
to the current centroids of the clusters, and finally the obtained clusters are
separated by a PWL function.

\subsection{Contribution}
This paper proposes a general supervised learning method for regression and/or 
classification of 
multiple targets that results in a PWA predictor over a single PWA partition of 
the feature space
in $K$ polyhedral cells. In each polyhedron, the predictor is either affine 
(for numeric targets) or given by the 
max of affine functions, i.e., convex piecewise 
affine (for categorical targets). Our 
goal is to obtain an overall predictor
that admits a simple encoding with binary and real variables, to be able
to solve optimization problems involving the prediction function via 
mixed-integer linear or quadratic 
programming. The number $K$ of 
linear predictors 
is therefore limited by the tolerated complexity of the resulting 
mixed-integer encoding
of the PWA predictor.

Rather than first clustering the training data 
and fitting $K$ different linear predictors, and then finding a PWL separation 
function to get the PWA 
partition, we simultaneously cluster, PWL-separate, and fit
by solving
a block-coordinate descent problem, similarly to the K-means 
algorithm~\citep{Llo57},
where we alternate between fitting models/separating clusters and reassigning 
training data
to clusters. We call the algorithm PARC (Piecewise Affine Regression and 
Classification) and show
that it converges in a finite number of iterations by showing that the sum of 
the loss functions 
associated with regression, classification, piecewise linear 
separation errors
decreases at each iteration. PWL separation is obtained by solving softmax 
regression problems or,
as a simpler alternative, by taking the Voronoi partition induced by the 
cluster centroids.

We test the PARC algorithm on different synthetic and real-world datasets.
After showing that PARC can reconstruct an underlying PWA function from its 
samples,
we investigate the effect of $K$ in reconstructing a nonlinear function, also 
showing how
to optimize with respect to the feature  vector so that the corresponding 
target is as close
as possible to a given reference value. Then we test PARC on many real-world 
datasets proposed 
for regression and classification, comparing its performance to alternative 
regression and classification 
techniques that admit a mixed-integer encoding of the predictor
of similar complexity, such as simple
neural networks based on ReLU activation functions and small decision trees.

A Python implementation of the PARC algorithm is available at 
\url{http://cse.lab.imtlucca.it/~bemporad/parc}.

\subsection{Outline}
After formulating the multivariate PWL regression and 
classification problem in 
Section~\ref{sec:statement}, we describe the proposed PARC algorithm and prove
its convergence properties in Section~\ref{sec:PARC}. In 
Section~\ref{sec:predictor} we
define the PWA prediction function for regression and classification, showing
how to encode it
using mixed-integer linear inequalities using big-M techniques. 
Section~\ref{sec:examples}
presents numerical tests on synthetic and real-world datasets. Some 
conclusions are
finally drawn in Section~\ref{sec:conclusions}.

\subsection{Notation and definitions}
Given a finite set $\CC$, $\card \CC$ denotes its number of elements
(cardinality). Given a vector $a\in\rr^n$, $\|a\|_2$ is the Euclidean norm of 
$a$,
$[a]_i$ denotes the $i$th component of $a$.
Given two vectors $a,b\in\rr^n$, we denote by
$[a=b]$ the binary quantity that is $1$ if $a=b$ or $0$ otherwise.
Given a matrix $A\in\rr^{m\times n}$,
$\|A\|_F$ denotes the Frobenius norm of $A$.
Given a polyhedron $P\subseteq\rr^n$, $\mathring{P}$ denotes
its interior. Given a finite set $S$ of real numbers $\{s_1,\ldots,s_K\}$
we denote by 
\begin{equation}
    \arg\min_{s\in S}=\min_{h}\{h\in\{1,\ldots,K\}: s_h\leq s_j,\ \forall j\in\{1,\ldots,K\}\}
    \label{eq:argmin}
\end{equation}
Taking the smallest index $h$ in~\eqref{eq:argmin} breaks ties in case of 
multiple minimizers.
The $\arg\max$ function of a set $S$ is defined similarly by replacing 
$ s_h\leq s_j$ with  $s_h\geq s_j$ in~\eqref{eq:argmin}.

\begin{definition}
A collection $\PP$ of sets $\{P_1,\ldots,P_K\}$ is said a \emph{polyhedral partition} of $\rr^n$ if $P_i$ is a polyhedron, $P_i\subseteq\rr^n$,
    $\forall i=1,\ldots,K$, $\cup_{i=1}^K P_i=\rr^n$, and $\mathring{P}_i\cap\mathring{P}_j=\emptyset$, $\forall i,j=1,\ldots,K$, $i\neq j$.
\end{definition}

\begin{definition}
    \label{def:PWA}
    A function $j:\rr^n\rightarrow\{1,\ldots,K\}$ is said \emph{integer 
    piecewise constant} (IPWC)~\citep{CB17} if there exist a polyhedral 
    partition 
    $\PP=\{P_1,\ldots,P_K\}$ of $\rr^n$ such that
    \begin{equation}
        j(x) = \arg\min_{h} \{h\in\{1,\ldots,K\}:\ x\in P_h\}
        \label{eq:ipwc-fun}
    \end{equation}
    for all $x\in\rr^n$.
\end{definition}
The ``$\arg\min$'' in~\eqref{eq:ipwc-fun} prevents possible
multiple definitions of $j(x)$ on overlapping boundaries $P_i\cap P_j\neq 
\emptyset$.
\begin{definition}
    \label{def:PWA2}
    A function $f:\rr^n\to\rr^m$ is said \emph{piecewise affine} (PWA) if there exists an IPWC function $j:\rr^n\to\{1,\ldots,K\}$ defined over a polyhedral partition $\PP$ and $K$ pairs $(a^i,b^i)$, $a^i\in\rr^{m\times n}$,
    $b^i\in\rr^m$, such that 
    \begin{equation}
        f(x) = a^{j(x)}x+b^{j(x)}
        \label{eq:pwa-fun}
    \end{equation}
    for all $x\in\rr^n$.
    It is said \emph{piecewise constant} if $a^i=0$, $\forall i\in\{1,\ldots,K\}$.
\end{definition}

\begin{definition}
    A \emph{piecewise linear (PWL) separation} function 
    $\Phi:\rr^n\to\rr$~\citep{BM94}
    is defined by
\begin{subequations}
    \beqar
    \Phi(x)&=&\omega^{j(x)}x+\gamma^{j(x)}\label{eq:PWL-sep-a}\\
    j(x)&=&\displaystyle{\min\left\{\arg\max_{j=1,\ldots,K}\{\omega^jx+\gamma^j\}\right\}}
    \label{eq:PWL-sep-b}
    \eeqar
    \label{eq:PWL-sep}%
\end{subequations}
where $\omega^j\in\rr^n$, $\gamma^j\in\rr$, $\forall j=1,\ldots,K$.
\end{definition}
A PWL separation function is convex~\citep{Sch87} and PWA
over the polyhedral partition $\PP=\{P_1,\ldots,P_K\}$ where
\begin{equation}
    P_j=\{x\in\rr^n:\ (\omega^h-\omega^j)x\leq \gamma^j-\gamma^h,\ \forall 
    h=1,\ldots,K,\ h\neq j\},\ j=1,\ldots,K
    \label{eq:PWL-partition}
\end{equation}

\section{Problem statement}
\label{sec:statement}
We have a training dataset $(x_k,y_k)$, $k=1,\ldots,N$, 
where $x_k$ contains $n_c$ numerical and $n_d$ categorical
features, each one of the latter containing $n_i$ possible values 
$\{v^i_1,\ldots,v^i_{n_i}\}$,
$i=1,\ldots,n_d$, and $y_k$ contains $m_c$ numerical targets and $m_d$ categorical targets,
each one containing $m_i$ possible values $\{w^i_1,\ldots,w^i_{m_i}\}$, 
$i=1,\ldots,m_d$. We assume that categorical features 
have been one-hot encoded into $n_i-1$ binary values,
so that $x_k\in\XX$, $\XX=\rr^{n_c}\times\{0,1\}^{s_x}$, 
$s_x=\sum_{i=1}^{n_d}(n_i-1)$.
By letting $n=n_c+s_x$ we have $x_k\in\rr^n$. Moreover, let 
$y_k=\smallmat{y_{ck}\\y_{dk}}$,
$y_{ck}\in\rr^{m_c}$, $[y_{dk}]_i\in\{w^i_1,\ldots,w^i_{m_i}\}$,
$\forall i=1,\ldots,m_d$, and define
$\YY=\rr^{m_c}\times
\{w^1_1,\ldots,w^1_{m_1}\}\times\ldots\times\{w^{m_d}_1,\ldots,w^{m_d}_{m_{m_d}}\}$,
so that we have $y_k\in\YY$.

Several approaches exist to solve regression problems to predict the numerical
components $y_c$ and classification problems for the categorical target vector $y_d$.
In this paper, we are interested in generalizing linear predictors for
regression and classification to \emph{piecewise linear} predictors $\hat 
y:\rr^n\to\YY$ over a single \emph{polyhedral partition} 
$\PP=\{P_1,\ldots,P_{K}\}$ of 
$\rr^n$.
More precisely, we want to solve the posed multivariate regression and 
classification problem by finding the following predictors
\begin{subequations}
    \beqar
    [\hat y_c(x)]_i&=&a^{j(x)}_ix+b^{j(x)}_i,\ i=1,\ldots,m_c\label{eq:PWA-regression}\\\null    
    [\hat y_d(x)]_i&=&w^i_{h},\ h=\arg\max_{t\in I(i)}\{a^{j(x)}_tx+b^{j(x)}_t\},\ i=1,\ldots,m_d
    \label{eq:PWA-classifier}%
    \eeqar%
    \label{eq:PWA-predictors}%
\end{subequations}
where $j(x)$ is defined as in~\eqref{eq:ipwc-fun} and
the coefficient/intercept values $a^j\in\rr^n$, $b^j\in\rr$ 
define a PWA function $f:\rr^n\to\rr^m$ as in~\eqref{eq:pwa-fun},
in which $m=m_c+\sum_{i=1}^{m_d}m_i$. In~\eqref{eq:PWA-predictors}, $I(i)$ 
denotes 
the set of 
indices 
corresponding to the $i$th categorical target $[y_d]_i$,
$I(i)=\{t(i)+1,\ldots,t(i)+m_i\}$, $t(i)=m_c+\sum_{h=1}^{i-1}m_h$.
Note that subtracting the same quantity $\bar ax+\bar b$
from all the affine terms in~\eqref{eq:PWA-classifier} does not change the maximizer,
for any arbitrary $\bar a\in\rr^n$, $\bar b\in\rr$.
To well-pose $\hat y_d$, according to~\eqref{eq:argmin} we also assume that 
the smallest index is taken in case ties occur when taking the maximum
in~\eqref{eq:PWA-classifier}.

We emphasize that all the components of $\hat y(x)$ in~\eqref{eq:PWA-predictors}
share the \emph{same polyhedral partition} $\PP$.
A motivation for this requirement is to be able
to efficiently solve optimization problems involving the resulting predictor $\hat y$
using mixed-integer programming, as we will detail in Section~\ref{sec:MIP}. 
Clearly, if this is not a requirement, by treating each 
target independently the problem can be decomposed in $m_c$ 
PWA regression problems and $m_d$ PWA classification problems.

Our goal is to jointly separate the training dataset in
$K$ clusters $\CC_1,\ldots,\CC_K$, $\CC_1=\{x_k: k\in J_j\}$,
where $\cup_{i=1}^KJ_i=\{1,\ldots,N\}$, $J_i\cap J_j=\emptyset$, $\forall i,j\in\{1,\ldots,N\}$, $i\neq j$, and to find optimal coefficients/intercepts
$a^j$, $b^j$ for~\eqref{eq:PWA-predictors}. In particular, if the clusters
were given, for each numerical target $[y_c]_i$, $i=1,\ldots,m_c$, we solve the 
ridge regression problem
\begin{equation}
    \min_{a^j_i,b^j_i} \alpha_j(\|a^j_i\|_2^2+(b^j_i)^2)+\sum_{k\in J_j} (y_{ki}-a^j_ix_k-b^j_i)^2
    \label{eq:ridge}
\end{equation}
with respect to the vector $a^j_i\in\rr^n$ of coefficients and intercept $b^j_i\in\rr$,
where $\alpha_j=\frac{\card{J_j}}{N}\alpha$ and $\alpha>0$
is an $\ell_2$-regularization parameter. For each binary target $[y_d]_i$, 
$i=1,\ldots,m_d$,
we solve the regularized softmax regression problem, a.k.a. Multinomial 
Logistic Regression (MLR) problems~\citep{Cox66,Thi69},
\begin{eqnarray}
    \min_{\scriptsize\ba{c}\{a^j_h,b^j_h\}\\h\in I(i)\ea}&&
    \sum_{h\in I(i)}\alpha_j(\|a^j_h\|_2^2+(b^j_h)^2)-\sum_{h=1}^{m_i}
    \sum_{\scriptsize \ba{c}
        k\in J_j:\\\null
        [y_{dk}]_i=w^i_h\ea}
    \log\frac{e^{a^j_{h+t(i)} x_k+b^j_{h+t(i)}}}
    {\sum_{t\in I(i)} e^{a^j_tx_k+b^j_t}}
    \label{eq:softmax-regression}
\end{eqnarray}
Note that, by setting $\alpha>0$, both~\eqref{eq:ridge} and~\eqref{eq:softmax-regression}
are strictly convex problems, and therefore their optimizers are unique.
It is well known that in the case of binary targets $[y_d]_i\in\{0,1\}$, 
problem~\eqref{eq:softmax-regression} is equivalent to
the regularized logistic regression problem
\begin{equation}
    \min_{a^j_h,b^j_h} \alpha_j(\|a^j_h\|_2^2+(b^j_h)^2)+\sum_{k\in J_j} \log\left(1+e^{(1-2[y_{dk}]_i)(a^j_ix_k+b^j_i)}\right)        
    \label{eq:logistic}
\end{equation}
where $h=t(i)+1$. Similarly, for preparing the 
background for what will follow in the
next sections, we can rewrite~\eqref{eq:softmax-regression} as
\begin{eqnarray}
\min_{\scriptsize\ba{c}\{a^j_h,b^j_h\}\\h\in I(i)\ea}&&\hspace{-.2cm}
    \sum_{h\in I(i)}\alpha_j(\|a^j_h\|_2^2+(b^j_h)^2)+\sum_{h=1}^{m_i}
    \sum_{\scriptsize \ba{c}
        k\in J_j:\\\null
        [y_{dk}]_i=w^i_h\ea}
    \log\left(\sum_{t\in I(i)} e^{a^j_tx_k+b^j_t}\right)    
    \nonumber\\
    &&\hspace*{-1.5cm}-a^j_{h+t(i)}x_k-b^j_{h+t(i)}=\hspace*{-.2cm}\min_{\scriptsize\ba{c}\{a^j_h,b^j_h\}\\h\in
     
    I(i)\ea}\hspace{-.2cm}
    \sum_{h\in I(i)}\alpha_j(\|a^j_h\|_2^2+(b^j_h)^2)+\sum_{k\in J_j}
    \log\left(\sum_{t\in I(i)} e^{a^j_tx_k+b^j_t}\right)\nonumber\\
    &&\hspace*{-1.5cm}-\sum_{h=1}^{m_i}[[y_{dk}]_i=w^i_h](a^j_{h+t(i)}x_k+b^j_{h+t(i)})
    \label{eq:softmax-regression-2}
\end{eqnarray}

\subsection{Piecewise linear separation}
\label{sec:PWL-separation}
Clustering the feature vectors $\{x_k\}$ in $\CC_1,\ldots,\CC_K$ should be 
based on two goals. On the one hand, we wish to have all the data values 
$(x_k,y_k)$ that can be best predicted
by $(a^j,b^j)$  in the same
cluster $\CC_j$. On the other hand, we would like  the clusters $\CC_1,\ldots,\CC_K$ 
to be piecewise linearly separable, i.e., that there exist a PWL separation 
function
$\Phi:\rr^n\to\rr$ as in~\eqref{eq:PWL-sep} such that $\CC_i\subseteq P_i$.
The above goals are usually conflicting (unless $y_k$ is a piecewise 
linear function of $x_k$), and we will have to trade them off.

Several approaches exist to find a PWL separation function $\Phi$ of given clusters
$\CC_1,\ldots,\CC_K$, usually attempting at minimizing the number of misclassified feature vectors $x_k$ (i.e., $x_k\in\CC_i$ and $x_k\not\in P_i$) in case the clusters are not piecewise-linearly separable. 
Linear programming was proposed in~\citep{BM94} to solve the following problem
\[
    \min_{\omega,\gamma} \sum_{j=1}^K\sum_{\scriptsize\ba{c}h=1\\h\neq j\ea}^K
    \sum_{k:\ x_k\in\CC_j}^N \frac{1}{\card \CC_j}\max\{(\omega^h-\omega^j)x_k+\gamma^j-\gamma^j+1,0\}
\]
Other approaches based on the piecewise smooth optimization 
algorithm of~\citep{BBP15} and averaged stochastic gradient descent
\citep{Bot12} were described in \citep{BPB16a}. In this paper, we 
use instead $\ell_2$-regularized softmax regression 
\begin{subequations}
    \begin{equation}
    \ba{rl}
    \min_{\omega,\gamma} & 
    \beta(\|\omega\|_F^2+\|\gamma\|_2^2)+\displaystyle{\sum_{j=1}^K\sum_{k:\
     
    x_k\in\CC_j}
        -\log\frac{e^{\omega^jx_k+\gamma^j}}{\sum_{i=1}^K
            e^{\omega^ix_k+\gamma^i}}}
    \ea
    \label{eq:softmax-cost}
\end{equation}
with $\beta\geq 0$, whose solution $\omega,\gamma$ provides the PWL separation 
function~\eqref{eq:PWL-sep}
as
\begin{equation}
    j(x)=\arg\max_{j=1,\ldots,K}
\frac{e^{\omega^jx+\gamma^j}}{\sum_{i=1}^K
    e^{\omega^ix+\gamma^i}}=
\arg\max_{j=1,\ldots,L} \omega^jx+\gamma^j    
    \label{eq:PWL-softmax}
\end{equation}%
\label{eq:softmax}%
\end{subequations}
and hence a polyhedral partition $\PP$ of the 
feature vector space as in~\eqref{eq:PWL-partition}. Note that,
as observed earlier, there are infinitely many PWL functions $\Phi(x)$
as in~\eqref{eq:PWL-sep-a} providing the same piecewise-constant function 
$j(x)$.
Hence, as it is customary, one can set one pair $(\omega^i,\gamma^i)=(0,0)$,
for instance $\omega^K=0$, $\gamma^K=0$
(this is equivalent to dividing both the numerator and denominator in the first 
maximization in~\eqref{eq:PWL-softmax} by $e^{\omega^Kx+\gamma^K}$), and
solve the reduced problem
\begin{equation}
    \ba{rl}
    \min_{\{\omega^j,\gamma^j\}_{j=1}^{K-1}} 
    &\beta(\|\omega\|_F^2+\|\gamma\|_2^2)+ \displaystyle{\sum_{j=1}^K\sum_{k:\ 
    x_k\in\CC_j}
        -\log\frac{e^{\omega^jx_k+\gamma^j}}{1+\sum_{i=1}^{K-1}
            e^{\omega^ix_k+\gamma^i}}}
    \ea
    \label{eq:softmax-2}
\end{equation}

An alternative approach to softmax regression is to obtain $\PP$ from the 
Voronoi diagram of the centroids
\begin{equation}
    \bar x_j=
    \arg\min_{x}\sum_{k\in J_j}\|x_k-x\|_2^2
    =\frac{1}{\card{\CC_j}}\sum_{k\in J_j}x_k
    \label{eq:centroids}
\end{equation}
of the clusters, inducing the PWL separation function as in~\eqref{eq:PWL-sep}
with
\begin{subequations}
\begin{eqnarray}
     j(x)&=&\arg\min_{j=1,\ldots,K}\|x-\bar x_j\|_2^2=
\arg\max_{j=1,\ldots,K}\omega^j x+\gamma^j\\
\omega^j&=&\bar x_j',\ \gamma^j=-\frac{1}{2}\|\bar x_j\|_2^2
\end{eqnarray}
    \label{eq:Voronoi-partition}%
\end{subequations}
Note that the Voronoi partitioning~\eqref{eq:Voronoi-partition} has $Kn$ degrees of freedom (the centroids $\bar x_j$), while softmax regression~\eqref{eq:PWL-softmax} has $Kn+(K-n-1)$ degrees of freedom.

\section{Algorithm}
\label{sec:PARC}
In the previous section, we have seen how to get the coefficients $a^j,b^j$
by ridge~\eqref{eq:ridge} or softmax~\eqref{eq:softmax-regression} regression
when the clusters $\CC$ are given, and how to get a PWL partition of $\CC$.
The question remains on how to determine the clusters $\CC_1,\ldots,\CC_K$.

Let us assume that the coefficients $a^j,b^j$ have been fixed.
Following~\eqref{eq:ridge} and~\eqref{eq:softmax-regression-2}
we could assign each training vector $x_k$ to the corresponding cluster $\CC_{j}$ such that
the following weighted sum of losses
\begin{eqnarray}
    V^y(a^j,b^j,x_k,y_k)&=&\sum_{i=1}^{m_c}\mu_{ci}(y_{ki}-a^j_ix_k-b^j_i)^2\\
    &&\hspace{-3cm}+\sum_{i=1}^{m_d}\mu_{di}\log\left(\sum_{t\in I(i)} 
    e^{a^j_tx_k+b^j_t}\right)
    -\sum_{h=1}^{m_i}[[y_{dk}]_i=w^i_h](a^j_{h+t(i)}x_k+b^j_{h+t(i)})
    \label{eq:cost_k-y}
\end{eqnarray}
is minimized, where $\mu_c\in\rr^{m_c}$, $\mu_d\in\rr^{m_d}$ are
vectors of relative weights on fit losses.

Besides the average quality of prediction~\eqref{eq:cost_k-y}, we also want to 
consider the location of the feature vectors $x_k$
to promote PWL separability of the resulting clusters
using the two approaches (softmax regression and Voronoi diagrams)
proposed in Section~\ref{sec:PWL-separation}. Softmax regression induces the 
criterion
\begin{subequations}
\begin{equation}
    V^x_s(\omega^j,\gamma^j,x_k)=-
    \log\frac{e^{\omega^jx_k+\gamma^j}}{1+\sum_{i=1}^{K-1}
        e^{\omega^ix_k+\gamma^i}}=
    \log\left(1+\sum_{i=1}^{K-1}
    e^{\omega^ix_k+\gamma^i}\right)-\omega^jx_k-\gamma^j
    \label{eq:cost_k-x-2}
\end{equation}
for $j=1,\ldots,K$, 
where $\omega^K=0$, $\gamma^K=0$.
Note that the last logarithmic term in~\eqref{eq:cost_k-x-2} does not
depend on $j$, so that it might be neglected in case $V^x_s$ gets minimized
with respect to $j$.

Alternatively, because of~\eqref{eq:Voronoi-partition},
Voronoi diagrams suggest penalizing the distance between $x_k$ and the 
centroid $\bar x_j$ of the cluster
\begin{equation}
    V^x_v(\bar x_j,x_k)=\|x_k-\bar x_j\|_2^2
    \label{eq:cost_k-x-1}
\end{equation}
Criteria~\eqref{eq:cost_k-x-2} and~\eqref{eq:cost_k-x-1} can be
combined as follows:
\begin{equation}
    V^x(\omega^j,\gamma^j,x_k)=
    \left\{\ba{ll}
    V^x_s(\omega^j,\gamma^j,x_k) & \mbox{if PWL 
    partitioning~\eqref{eq:PWL-partition} is used}\\
        V^x_v((\omega^j)',x_k)+0\cdot\gamma_j & \mbox{if Voronoi 
        partitions~\eqref{eq:Voronoi-partition} are used}
    \ea\right.
\end{equation}
    \label{eq:cost_k}%
\end{subequations}
Then, each training vector $x_k$ is assigned to the cluster $\CC_{j_k}$ such that
\begin{equation}
    j_k=\arg\min_{j=1,\ldots,K} V^y(a^j,b^j,x_k,y_k)+\sigma V^x(\omega^j,\gamma^j,x_k)
    \label{eq:assignment}
\end{equation}
where $\sigma\geq 0$ is a relative weight that allows trading off between target fitting and
PWL separability of the clusters. Note that, according to the definition 
in~\eqref{eq:argmin}, in the case of multiple minima the optimizer $j_k$ 
in~\eqref{eq:assignment} is always selected as the smallest index among optimal 
indices. 

The idea described in this paper is to alternate between fitting linear 
predictors as in~\eqref{eq:ridge}--\eqref{eq:softmax-regression} and 
reassigning vectors 
to clusters as in~\eqref{eq:assignment}, as described in 
Algorithm~\ref{algo:PARC}
that we call PARC (Piecewise Affine Regression and Classification).

The following theorem proves that indeed PARC is an algorithm, as it terminates in a finite number of steps to a local minimum of the problem of finding the $K$ best linear predictors.
\begin{theorem}
    \label{th:convergence}
    Algorithm~\ref{algo:PARC} converges in a finite number of steps
    to a local minimum of the following mixed-integer optimization problem
    \begin{subequations}
        \beqar
        \min_{a,b,\omega,\gamma,z} && V(a,b,\omega,\gamma,z)\nonumber\\
        \st&& \displaystyle{\sum_{j=1}^Kz_{kj}=1},\ \forall k=1,\ldots,N
        \label{eq:PARC-optim}\\
        \hspace*{-1cm}V(a,b,\omega,\gamma,z)&=&\sigma\beta(\|\omega\|_F^2+\|\gamma\|_2^2)+
        \sum_{j=1}^K\sum_{k=1}^Nz_{kj}\left(
        \frac{\alpha}{N}(\|a^j\|_F^2+\|b^j\|_2^2)+\right.\nonumber\\
        &&   \left.
        V^y(a^j,b^j,x_k,y_k)+\sigma V^x(\omega^j,\gamma^j,x_k)\right)
        \label{eq:PARC-problem-cost}
        \eeqar
        \label{eq:PARC-problem}%
    \end{subequations}
    where $a^j\in\rr^{m\times n}$, $b^j\in\rr^{m}$,
    $\omega^j\in\rr^{K\times n}$,$\gamma^j\in\rr^{K}$, 
    $\forall j=1,\ldots,K$, $z\in\{0,1\}^{N\times K}$, and
    with either $\omega^K=0$, $\gamma^K=0$, and $\beta\geq 0$ if PWL 
    partioning~\eqref{eq:PWL-partition} is used, or 
    $\gamma^j=-\frac{1}{2}\|\omega^j\|_2^2$, $\forall j=1,\ldots,K$, and 
    $\beta=0$ 
    if Voronoi 
    partions~\eqref{eq:Voronoi-partition} are used.
\end{theorem}

\proof
We prove the theorem by showing that Algorithm~\ref{algo:PARC}
is a block-coordinate descent algorithm for problem~\eqref{eq:PARC-problem},
alternating between the minimization with respect to $(a,b,\omega,\gamma)$ and 
with respect to $z$. The proof follows arguments similar to those used to prove 
convergence of unsupervised learning approaches like K-means. 
The binary variables $z_{kj}$ are hidden variables such that $z_{kj}=1$ if and 
only if
the target vector $y_k$ is predicted by $j(x_k)=j$ as in~\eqref{eq:PWA-predictors}.

The initial clustering $\CC_1,\ldots,\CC_K$ of $\{x_k\}$ 
determines the initialization of the latent variables, i.e., $z_{kj}=1$ if and only if $x_k\in\CC_j$, or equivalently $k\in J_j$. Let us consider $z$ fixed. 
Since
\begin{eqnarray*}
   \sum_{j=1}^K\sum_{k=1}^Nz_{kj}\left(\frac{\alpha}{N}(\|a^j\|_F^2+\|b^j\|_2^2)\right)&
   =&\sum_{j=1}^K\frac{\card{J_j}}{N}\alpha(\|a^j\|_F^2+\|b^j\|_2^2)\\
   &=&\sum_{j=1}^K\sum_{i=1}^{m_c+m_d}\alpha_j(\|a^j_i\|_2^2+(b^j_i)^2)
\end{eqnarray*}
problem~\eqref{eq:PARC-problem} becomes separable into ($i$) $Km_c$ 
independent optimization problems
of the form~\eqref{eq:ridge}, ($ii$) $Km_d$ softmax regression problems as in~\eqref{eq:softmax-regression},
and ($iii$) either a softmax regression problem as in~\eqref{eq:softmax-cost}
or $K$ optimization problems as in~\eqref{eq:centroids}.

Let $a^j$, $b^j$, $\omega^j$, $\gamma^j$ be the solution to such problems and consider now them fixed. In this case, problem~\eqref{eq:PARC-problem} becomes
\begin{equation}
    \ba{rl}
    \min_{z\in\{0,1\}^{N\times K}} & \displaystyle{\sum_{k=1}^N\sum_{j=1}^Kz_{kj}\left(V^y(a^j,b^j,x_k,y_k)+\sigma V^x(a^j,b^j,\omega^j,\gamma^j,x_k)\right)}\\
    \st& \displaystyle{\sum_{j=1}^Kz_{kj}=1},\ \forall k=1,\ldots,N
    \ea
    \label{eq:all-assignments}
\end{equation}
which is separable with respect to $k$ into $N$ independent binary optimization problems.
The solution of~\eqref{eq:all-assignments} is given by computing $j_k$ as in~\eqref{eq:assignment} and by setting $z_{j_k}=1$ and $z_j=0$ for all $j=1,\ldots,K$,
$j\neq j_k$.

Having shown that PARC is a coordinate-descent
algorithm, the cost $V(a,b,\omega$, $\gamma$, $z)$ in~\eqref{eq:PARC-problem} 
is monotonically non-increasing at each iteration of Algorithm~\ref{algo:PARC}. 
Moreover,
since all the terms in the function are nonnegative, the sequence of optimal cost values is lower-bounded by zero, so it converges asymptotically.
Moreover, as the number of possible
combinations $\{z_{kj}\}$ are finite, Algorithm~\ref{algo:PARC} always terminates 
after a finite number of steps, since we have assumed that
the smallest index $j_k$ is always taken in~\eqref{eq:assignment}
in case of multiple optimizers. The latter implies that no 
chattering between different
combinations $z_{kj}$ having the same cost $V$ is possible.
\qed

\begin{algorithm}[t]
    \caption{PARC (Piecewise Affine Regression and Classification)}
    \label{algo:PARC}
    ~~\textbf{Input}: Training dataset $(x_k,y_k)$, $k=1,\ldots,N$; number
    $K$ of desired linear predictors; $\ell_2$-regularization parameters 
    $\alpha>0$, $\beta\geq 0$; fitting/separation tradeoff parameter 
    $\sigma\geq 0$;
    output weight vector $\mu\in\rr^m$, $\mu\geq 0$;
    initial clustering $\CC_1,\ldots,\CC_K$ of $\{x_k\}$.
    \vspace*{.1cm}\hrule\vspace*{.1cm}
    \begin{enumerate}[label*=\arabic*., ref=\theenumi{}]
        \item $i\leftarrow 1$; 
        \item \textbf{Repeat}
        \begin{enumerate}[label=\theenumi{}.\arabic*., ref=\theenumi{}.\arabic*]
            \item \textbf{For all} $j=1,\ldots,K$ \textbf{do} 
            \label{algo:PARC-for}           
            \begin{enumerate}[label=\theenumii{}.\arabic*., 
            ref=\theenumii{}.\arabic*]
                \item Solve the ridge regression problem~\eqref{eq:ridge}, 
                $\forall i=1,\ldots, m_c$;
                \label{algo:PARC-ridge}
                \item Solve the softmax regression 
                problem~\eqref{eq:softmax-regression}, $\forall i=m_c+1,\ldots, 
                m$;
                \label{algo:PARC-classification}
            \end{enumerate}
            \item PWL separation: either compute the cluster centroids 
            $\omega^j=\bar x_j'$~\eqref{eq:centroids} and set $\gamma_j=0$, 
            $j=1,\ldots,K$ (Voronoi partitioning), or 
            $\omega^j,\gamma^j$ as in~\eqref{eq:softmax-cost} (general PWL 
            separation);        \label{algo:PARC-separation}     
            \item \textbf{For all} $k=1,\ldots,N$ \textbf{do}            
            \label{algo:PARC-assign}
            \begin{enumerate}[label=\theenumii{}.\arabic*., 
            ref=\theenumii{}.\arabic*]
                \item Evaluate $j_k$ as in~\eqref{eq:assignment};
                \item Reassign $x_k$ to cluster $\CC_{j_k}$;
                \label{algo:PARC-assignment}
            \end{enumerate}
        \end{enumerate}
        \item \textbf{Until} convergence; \label{algo:PARC-until} 
        \item \textbf{End}.
    \end{enumerate}
    \vspace*{.1cm}\hrule\vspace*{.1cm}
    ~~\textbf{Output}: Final number $K_f$ of clusters; coefficients $a_j$ and 
    intercepts $b_j$ of linear functions, and $\omega^j,\gamma^j$ of PWL 
    separation function, $j=1,\ldots,K_f$, final clusters 
    $\CC_1,\ldots,\CC_{K_f}$.
\end{algorithm}

Theorem~\ref{th:convergence} proved that PARC converges
in a finite number of steps. Hence, a termination criterion
for Step~\ref{algo:PARC-until} of Algorithm~\ref{algo:PARC} 
is that $z$ does not change from the previous iteration.
An additional termination criterion is to introduce a tolerance $\epsilon>0$
and stop when the optimal cost $V(a,b,\omega,\gamma,z)$ has not decreased more than $\epsilon$
with respect to the previous iteration. In this case, as the reassignment
in Step~\ref{algo:PARC-assignment} may have changed the $z$ matrix,
Steps~\ref{algo:PARC-ridge}--\ref{algo:PARC-classification} must be executed
before stopping, in order to update the coefficients/intercepts $(a,b)$ 
accordingly.

Note that PARC is only guaranteed to converge to a local
minimum; whether this is also a global one depends on the provided initial 
clustering $\CC_1,\ldots,\CC_K$, i.e., on the initial guess on $z$.
In this paper, we initialize $z$ by running the K-means++ 
algorithm~\citep{AV07}
on the set of feature vectors $x_1,\ldots,x_N$. For solving single-target 
regression problems, an alternative approach to get the initial clustering
could be to associate to each datapoint  $x_k$ the coefficients $c_k$ of the 
linear 
hyperplane fitting 
the $K_n$ nearest neighbors of $x_k$ (cf.~\cite{FMLM03a}), 
for example, by setting $K_n=2(n+1)$, and then run K-means on the set 
$c_1,\ldots,c_n$ to get an 
assignment $\delta_k$. This latter approach, however, can be sensitive
to noise on measured targets and is not used in the numerical experiments
reported in Section~\ref{sec:examples}.

As in the K-means algorithm, some clusters may become empty
during the iterations, i.e., some indices $j$ are such that $z_{kj}=0$ for all $k=1,\ldots,N$. In this case, Step~\ref{algo:PARC-for} of
Algorithm~\ref{algo:PARC} only loops on the indices $j$ for which $z_{kj}=1$ 
for some $k$. Note that the values of $a^j$, $b^j$, $\omega^j$, and $\gamma^j$, 
where
$j$ is the index of an empty cluster, do not affect the value of the overall function $V$
as their contribution is multiplied by 0 for all $k=1,\ldots,N$. 
Note also that some categories may disappear from the subset of samples in the 
cluster in the case of multi-category targets. In this case, 
still~\eqref{eq:softmax-regression} provides
a solution for the coefficients $a_j^h,b^j_h$ corresponding to missing categories $h$, so
that $V^y$ in~\eqref{eq:cost_k-y} remains well posed. 

After the algorithm stops, clusters $\CC_j$ containing less than
$c_{\rm min}$ elements can be also eliminated, interpreting the corresponding 
samples as outliers (alternatively, their elements could be reassigned to the 
remaining clusters).
We mention that after the PARC algorithm terminates, for each numeric
target $[y_c]_i$ and cluster $\CC_j$ one can further fine-tune the 
corresponding 
coefficients/intercepts $a_i^j$,  $b_i^j$ by choosing the
$\ell_2$-regularization parameter 
$\alpha^j$ in each region via leave-one-out cross-validation on the subset of 
datapoints contained in the cluster. In case some features or targets have very 
different ranges, the numeric components in $x_k$, $y_k$ should be scaled.

Note that purely solving $m_c$ ridge and $m_d$ softmax regression on the entire dataset
corresponds to the special case of running PARC with $K=1$. Note also that, when $\sigma\rightarrow +\infty$, PARC will determine a PWL separation of the feature vectors, 
then solve $m_c$ ridge and $m_d$ softmax regression on each cluster. In this 
case, if the initial clustering $\CC$ is determined by K-means,
PARC stops after one iteration.

We remark that evaluating~\eqref{eq:cost_k-y} and \eqref{eq:cost_k-x-2} (as 
well as solving softmax regression problems) requires computing the logarithm 
of the sum of exponential, see, e.g., the recent paper 
\citep{BHH19} for numerically accurate implementations.

When the PWL separation~\eqref{eq:softmax-cost} is used, or in case of 
classification problems, most of the computation effort spent 
by PARC is due to solving softmax regression problems. In our implementation, 
we have used the general L-BFGS-B 
algorithm~\citep{BLBZ95},
with warm-start equal to the value obtained from the previous PARC iteration
for the same set of optimization variables. 
Other efficient methods for solving MLR problems have been proposed in the 
literature, such
as iteratively reweighted least squares (IRLS), that is a Newton-Raphson 
method~\citep{Lea90},
stochastic average gradient (SAG) descent~\citep{SLB17}, the alternating 
direction method of multipliers 
(ADMM)~\citep{BPCPE11}, and methods based on majorization-minimization 
(MM) methods~\citep{KCFH90,FSS15,JB20}.

We remark that PARC converges even if the softmax regression 
problem~\eqref{eq:softmax-cost} is not solved to optimality. Indeed, the proof 
of Theorem~\ref{th:convergence} still holds as long as the optimal cost
in~\eqref{eq:softmax-cost} decreases with respect to the last computed
value of $\omega,\gamma$. This suggests that during intermediate PARC
iterations, in case general PWL separation is used, to save computations
one can avoid using tight optimization tolerances in
Step~\ref{algo:PARC-separation}. Clearly, loosening
the solution of problem~\eqref{eq:softmax-cost} can impact 
the total number of PARC iterations; hence, there is a tradeoff to take into 
account.

We finally remark that Steps~\ref{algo:PARC-for} and~\ref{algo:PARC-assign} can 
be parallelized for speeding computations up.

\section{Predictor}
\label{sec:predictor}
After determining the coefficients $a^j$, $b^j$ by running PARC,
we can define the prediction functions $\hat y_c$, $\hat y_d$,
and hence the overall predictor $\hat y$ as in~\eqref{eq:PWA-predictors}.
This clearly requires defining $j(x)$,
i.e., a function that associates to any vector $x\in\rr^n$ the 
corresponding predictor out of the $K$ available. Note that the obtained 
clusters $\CC_j$ may not be piecewise-linearly separable. 

In principle any classification method on the dataset $\{x_k,\delta_k\}$,
where $\delta_k=j$ if and only if $x_k\in\CC_j$, can be used
to define $j(x)$. For example, nearest neighbors 
($j(x)=\arg\min_{k=1,\ldots,N}\|x-x_k\|_2^2$), decision trees, na\"ive Bayes, 
or one-to-all neural or support vector classifiers to mention a few. 
In this paper, we are interested in defining $j(x)$ using
a polyhedral partition $\PP=\{P_1,\ldots,P_K\}$ as stated in Section~\ref{sec:statement}, 
that is to select $j(x)$ such that it is IPWC as defined in~\eqref{eq:ipwc-fun}. 
Therefore, the natural choice is to use the values of $(\omega^j,\gamma^j)$ 
returned by PARC to define a PWL separation function 
by setting $j(x)$ as in~\eqref{eq:PWL-softmax}, which defines $P_j$ as 
in~\eqref{eq:PWL-partition}, or, if Voronoi partitioning is used in PARC, 
set $j(x)=\arg\min_{j=1,\ldots,K_f}\|x-\bar x_j\|_2^2$, which leads to 
polyhedral cells $P_j$ as in~\eqref{eq:Voronoi-partition}.
As the clusters $\CC_1,\ldots,\CC_{K_f}$ may not be piecewise-linearly 
separable, after defining the partition $\PP=\{P_1,\ldots,P_{K_f}\}$, one can 
cluster the datapoints again by redefining $\CC_j=\{x_k:\ x_k\in P_j,\ 
k=1,\ldots,N\}$ and then execute one last time 
Steps~\ref{algo:PARC-ridge}--\ref{algo:PARC-classification} of the PARC 
algorithm to get the final coefficients $a,b$ defining the predictors $\hat 
y_c$, $\hat y_d$.
Note that these may not be continuous functions of the feature vector $x$.

Finally, we remark that the number of floating point operations 
(flops) required to evaluate the predictor $\hat y(x)$ at a given $x$
is roughly $K$ times that of a linear predictor, as it involves  $K$ 
scalar products $[\omega^j\ \gamma^j]\smallmat{x\\ 1}$ as 
in~\eqref{eq:PWL-softmax} or~\eqref{eq:Voronoi-partition} ($2K(n_x+1)$ flops), 
taking their maximum, and then evaluate a 
linear predictor (another $2(n_x+1)$ flops per target in case of 
regression~\eqref{eq:PWA-regression} and $2m_i(n_x+1)$ flops and a maximum for 
multi-category targets~\eqref{eq:PWA-classifier}).

\subsection{Mixed-integer encoding}
\label{sec:MIP}
To optimize over the estimated model $\hat y$ we need to suitably encode
its numeric components $\hat y_c$ and categorical components $\hat y_d$ by introducing
binary variables. First, let us introduce a binary vector $\delta\in\{0,1\}^K$ to encode the PWL partition induced by~\eqref{eq:PWL-sep}
\begin{subequations}
\begin{eqnarray}
    \hspace*{-.5cm}(\omega^i-\omega^j)x&\leq& 
    \gamma^j-\gamma^i+M_{ji}(1-\delta_j),\
    \forall i=1,\ldots,K,\ i\neq j,\ \forall j=1,\ldots,K
    \label{eq:mip-pwa-bigM}\\
    \sum_{j=1}^K\delta_j&=&1\label{eq:mip-pwa-xor}
\end{eqnarray}
\label{eq:mip-partition}%
\end{subequations}
where $\omega^j,\gamma^j$ are the coefficients optimized by the PARC
algorithm when PWL separation~\eqref{eq:PWL-sep} is used (with
$\omega^K=0$, $\gamma^K=0$), or $\omega^j=\bar x_j'$ and 
$\gamma=-\|\bar x_j\|_2^2$ if Voronoi partitioning~\eqref{eq:Voronoi-partition} is used instead.
The constraint~\eqref{eq:mip-pwa-bigM} is the  ``big-M'' reformulation
of the logical constraint $[\delta_{j}=1]\rightarrow[x\in P_i]$,
that, together with the exclusive-or (SOS-1) constraint~\eqref{eq:mip-pwa-xor}
models the constraint $[\delta_{j}=1]\leftrightarrow[x\in P_i]$.
The values of $M_{ji}$ are upper-bounds that need to satisfy
\begin{equation}
    M_{ji}\geq \max_{x\in\BB} (\omega^i-\omega^j)x-\gamma^j+\gamma^i,\    
    \forall i,j=1,\ldots,K,\ i\neq j
    \label{eq:Mij}
\end{equation}
where $\BB\subset\rr^n$ is a compact subset of features of interest.
For example, given the dataset $\{x_k\}_{k=1}^N$ of features, we can
set $\BB$ as a box containing all the sample feature vectors
so that the values $M_{ij}$ in~\eqref{eq:Mij} can be easily 
computed by solving $K(K-1)$ linear programs. 
A simpler way to estimate the values $M_{ji}$ is given by the following
lemma \citep[Lemma 1]{LK00}:
\begin{lemma}
    \label{lemma:bigM}
    Let $\BB=\{x\in\rr^n:\ x_{\rm min}\leq x\leq x_{\rm max}$
    and $v\in\rr^n$. Let $v^+=\max\{v,0\}$, $v^-=\max\{v,0\}$. Then
    \begin{equation}
    \sum_{i=1}^n v_i^+x_{\rm min,i}-v_i^-x_{\rm max,i}\leq
    v'x\leq \sum_{i=1}^n v_i^+x_{\rm max,i}-v_i^-x_{\rm min,i}
        \label{eq:big-M-lemma}
    \end{equation}
\end{lemma}
\proof Since $x_{\rm min,i}\leq x_i\leq x_{\rm max,i}$ and $v=v^+-v^-$, we
get
\[
    v'x=\sum_{i=1}^nv_ix_i=\sum_{i=1}^n (v_i^+-v_i^-)x_i\leq
     \sum_{i=1}^n v_i^+x_{\rm max,i}-v_i^-x_{\rm min,i}
\]
and similarly $v'x\geq 
\sum_{i=1}^n v_i^+x_{\rm min,i}-v_i^-x_{\rm max,i}$.
\qed
By applying Lemma~\ref{lemma:bigM} for $v=\omega^i-\omega^j$,~\eqref{eq:Mij} is 
satisfied by setting
\begin{equation}
    M_{ji}=\gamma^i-\gamma^j+\sum_{h=1}^n \max\{\omega^i_h-\omega^j_h,0\}
    x_{\rm max,h}-\max\{\omega^j_h-\omega^i_h,0\}x_{\rm min,h}
    \label{eq:big-Mij-lemma}
\end{equation}
for all $i,j=1,\ldots,K,\ i\neq j$.

Having encoded the PWL partition, the $i$th predictor is given by
\begin{subequations}
\begin{equation}
    [\hat y_c(x)]_i=\sum_{j=1}^K p_{ij}
    \label{eq:ychat-1}
\end{equation}
where $p_{ij}\in\rr$ are optimization variables representing
the product $p_{ji}=\delta_j(a^j_ix+b_i^j)$. This is modeled by
the following mixed-integer linear inequalities
\begin{equation}
        \ba{rcl}
        p_{ji}&\leq&a^j_ix+b_i^j-M^{c-}_{ji}(1-\delta_j)\\
        p_{ji}&\geq&a^j_ix+b_i^j-M^{c+}_{ji}(1-\delta_j)\\
        p_{ji}&\leq&M^{c+}_{ji}\delta_j\\
        p_{ji}&\geq&M^{c-}_{ji}\delta_j 
        \ea       
    \label{eq:ychat-2}
\end{equation}
The coefficients $M^{c-}_{ji}$, $M^{c+}_{ji}$ need to satisfy
$M^{c-}_{ji}\leq \min_{x\in\BB}a^j_ix+b_i^j\leq
    \max_{x\in\BB}a^j_ix+b_i^j\leq M^{c+}_{ji}$
and can be obtained by linear programming or, more simply, by 
applying Lemma~\ref{lemma:bigM}.

Regarding the $m_d$ classifiers $\hat y_{di}$, to model the ``$\arg\max$'' 
in~\eqref{eq:PWA-classifier} we further introduce
$s_y$ binary variables $\nu_{ih}\in\{0,1\}$, $h=1,\ldots,m_i$, 
$i=1,\ldots,m_d$, satisfying the following big-M constraints
\begin{eqnarray}
&&\hspace*{-1.5cm}(a^j_h-a^j_t)x\geq 
b^j_t-b^j_h-M^d_{ht}(2-\nu_{ih}-\delta_j),\ 
\forall h,t\in I(i),\ h\neq t,\ \forall j=1,\ldots,K\label{eq:bigM-max}\\
&&\hspace*{-1.5cm}\sum_{h=1}^{m_i}\nu_{ih}=1,\ \forall i=1,\ldots,m_d
\label{eq:ydhat}
\end{eqnarray}
where the coefficients $M^d_{ht}$ must satisfy $\displaystyle{
    M^{d}_{ht}\geq 
    \max_{j=1,\ldots,K}\{\max_{x\in\BB}(a^j_t-a^j_h)x+b^j_t-b^j_h\}}
$. 
Note that the constraints in~\eqref{eq:bigM-max} become redundant when 
$\delta_j=0$ or $\nu_{ih}=0$
and lead to $a^j_hx+b^j_h\geq a^j_t+b^j_t$ for all $t\in I(i)$, $t\neq h$, when 
$\nu_{ih}=1$ and $\delta_j=1$, which is the binary equivalent of $[\hat 
y_d(x)]_i=w^i_h$ for $x\in P_j$.
Then, the $i$th classifier is given by
\begin{equation}
    [\hat y_d(x)]_i=\sum_{h=1}^{m_i}w^i_{h}\nu_{ih}
    \label{eq:ydhat-3}
\end{equation}
\label{eq:yhat-bigM}%
\end{subequations}
In conclusion,~\eqref{eq:mip-partition} and~\eqref{eq:yhat-bigM} provide 
a mixed-integer linear reformulation of the predictors $\hat y_c,\hat y_d$ 
as in~\eqref{eq:PWA-predictors} returned by the PARC algorithm. This enables
solving optimization problems involving the estimated model, possibly under
linear and logical constraints on features and targets. For example,
given a target vector $y_{\rm ref}$, the problem of 
finding the feature vector $x^*$ such that $\hat y_c(x^*)\approx y_{\rm ref}$
can be solved by minimizing $\|\hat y_c(x)-y_{\rm ref}\|_\infty$
as in the following mixed-integer linear program (MILP)
\begin{eqnarray}
    \min_{x,p,\delta,\epsilon}&&\epsilon\nonumber\\
    \st && \epsilon\geq \pm \left(\sum_{j=1}^K p_{ij}-y_{{\rm ref},i}\right)
            \label{eq:MILP-tracking}\\
    &&\mbox{Constraints~\eqref{eq:mip-partition},~\eqref{eq:ychat-1},~\eqref{eq:ychat-2}}
    \nonumber
\end{eqnarray}
The benefit of the MILP formulation~\eqref{eq:MILP-tracking} is that it
can be solved to global optimality by very efficient solvers. 
Note that if a more refined nonlinear
predictor $\hat y_{NL}$ is available, for example, a feedforward neural network
trained on the same dataset, the solution $x^*$ can be used to
warm-start a nonlinear programming solver based on $\hat y_{NL}$,
which would give better chances to find a global minimizer.

\section{Examples}
\label{sec:examples}
We test the PARC algorithm on different examples. First, we consider 
synthetic data generated from sampling a piecewise affine function and see 
whether PARC can recover the function. Second, we consider synthetic 
data from a toy example in which a nonlinear function generates the data, so to 
test the effect of the main hyper-parameters of PARC, namely $K$ and $\sigma$, 
also optimizing over the model using mixed-integer linear programming. In 
Section~\ref{sec:real-datasets} we will instead test PARC on several regression 
and classification examples on real datasets from the PMLB 
repository~\citep{Olson2017PMLB}.
All the results have been obtained in Python 3.8.3 on an Intel Core i9-10885H 
CPU @2.40GHz machine.
The \textsf{scikit-learn} package~\citep{scikit-learn} is used to solve
ridge and softmax regression problems, the latter using L-BFGS to solve the 
nonlinear programming problem~\eqref{eq:softmax-regression}.

\subsection{Synthetic datasets}
\label{sec:toy-datasets}

\subsubsection{Piecewise affine function}
\begin{figure}[th!]
    \centerline{
        \begin{tabular}[t]{c} 
            \subfigure[True PWA function~\eqref{eq:PWA-example-fun} and 
            dataset]{\parbox{.45\hsize}{\includegraphics[width=\hsize]{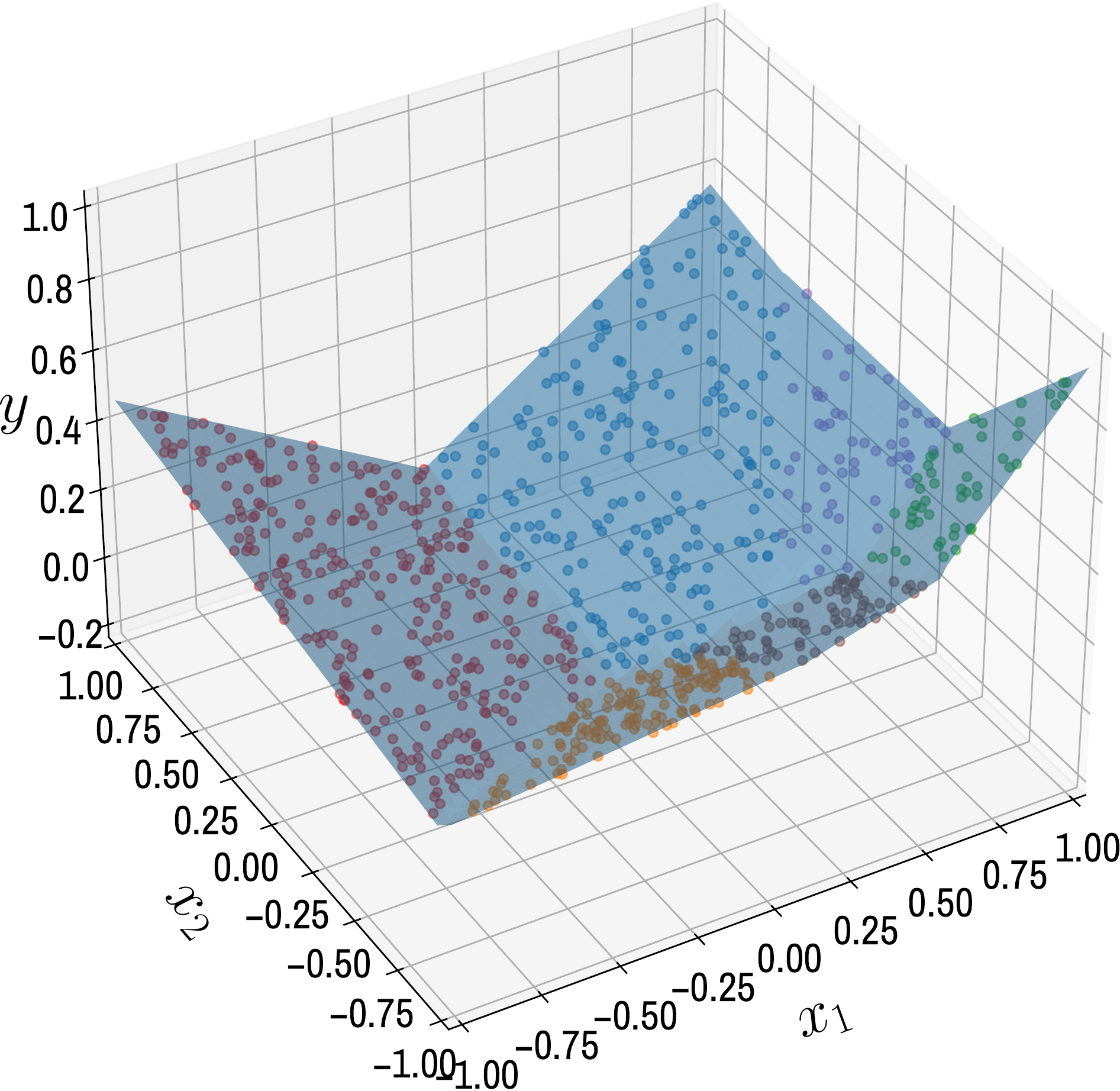}\\~}
                \label{fig:PWA_function_3}}
            \hspace{.1\hsize}
            \subfigure[Cost function during PARC iterations]       
            {\parbox{.45\hsize}{\includegraphics[width=\hsize]{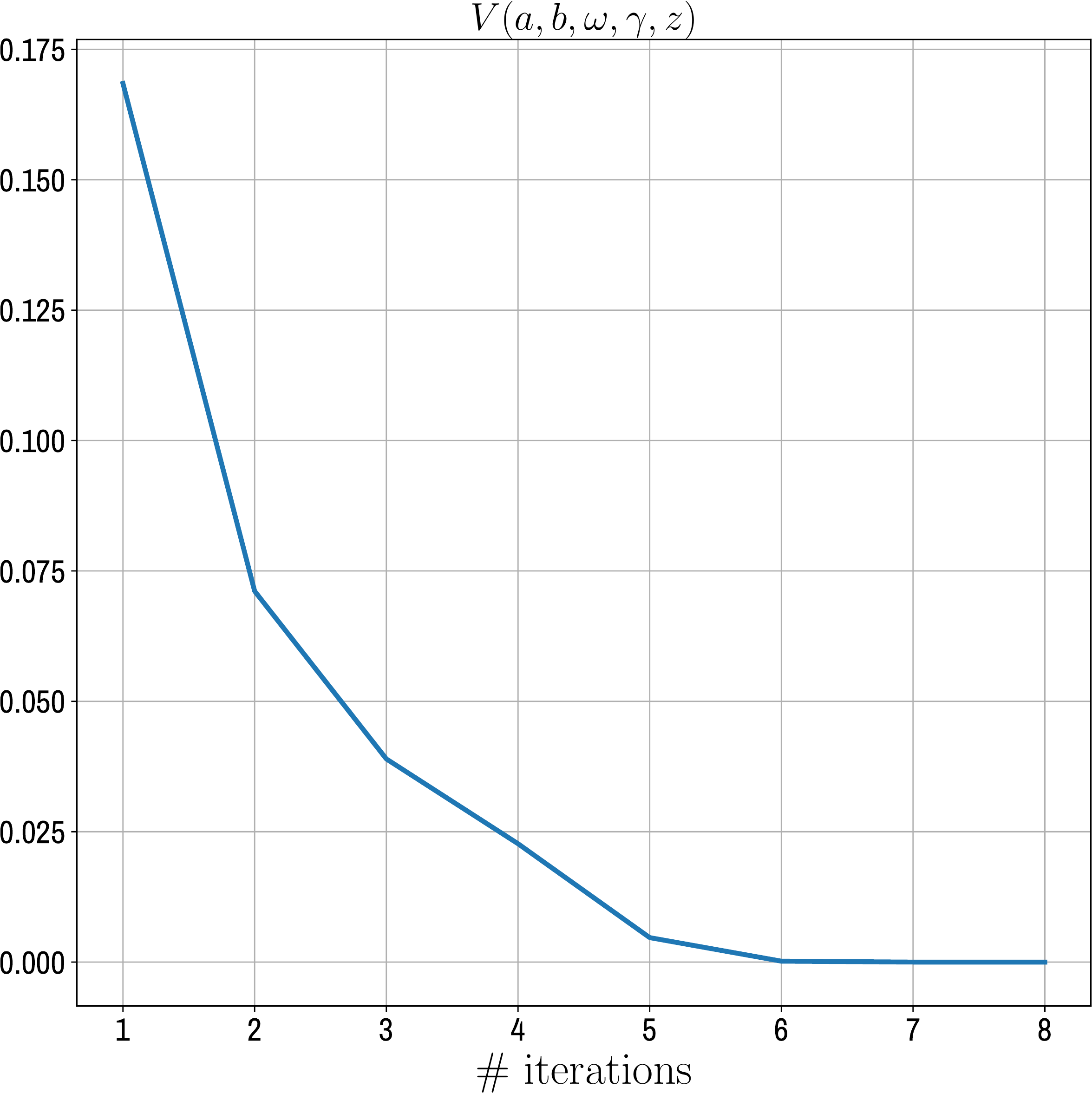}\\~}
                \label{fig:PWA_function_4}}\\
            \subfigure[True PWA partition induced 
            by~\eqref{eq:PWA-example-fun}]{\parbox{.45\hsize}{\includegraphics[width=\hsize]{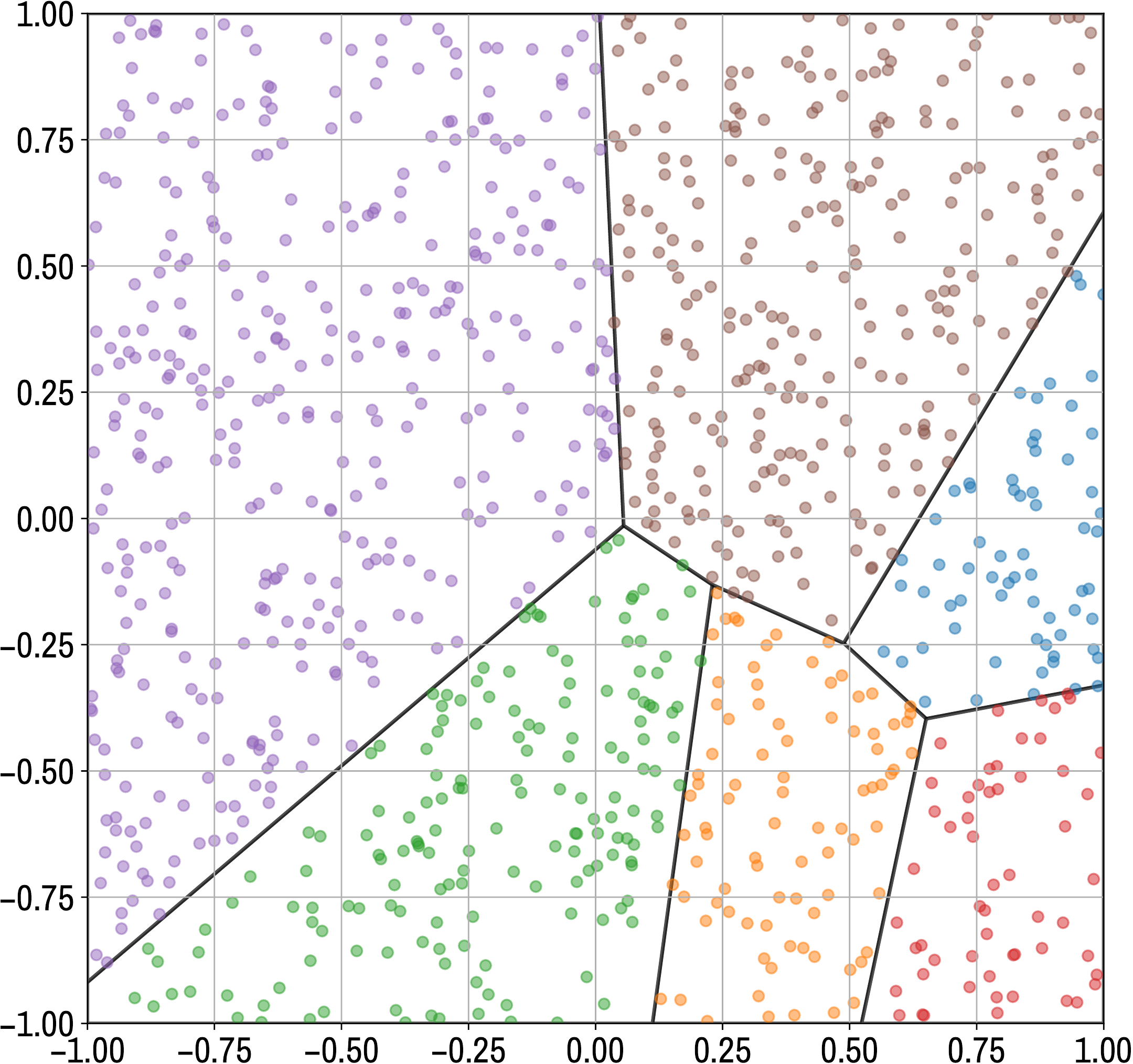}\\~}
                \label{fig:PWA_function_1}}
            \hspace{.1\hsize}
            \subfigure[PWA partition generated by PARC]       
            {\parbox{.45\hsize}{\includegraphics[width=\hsize]{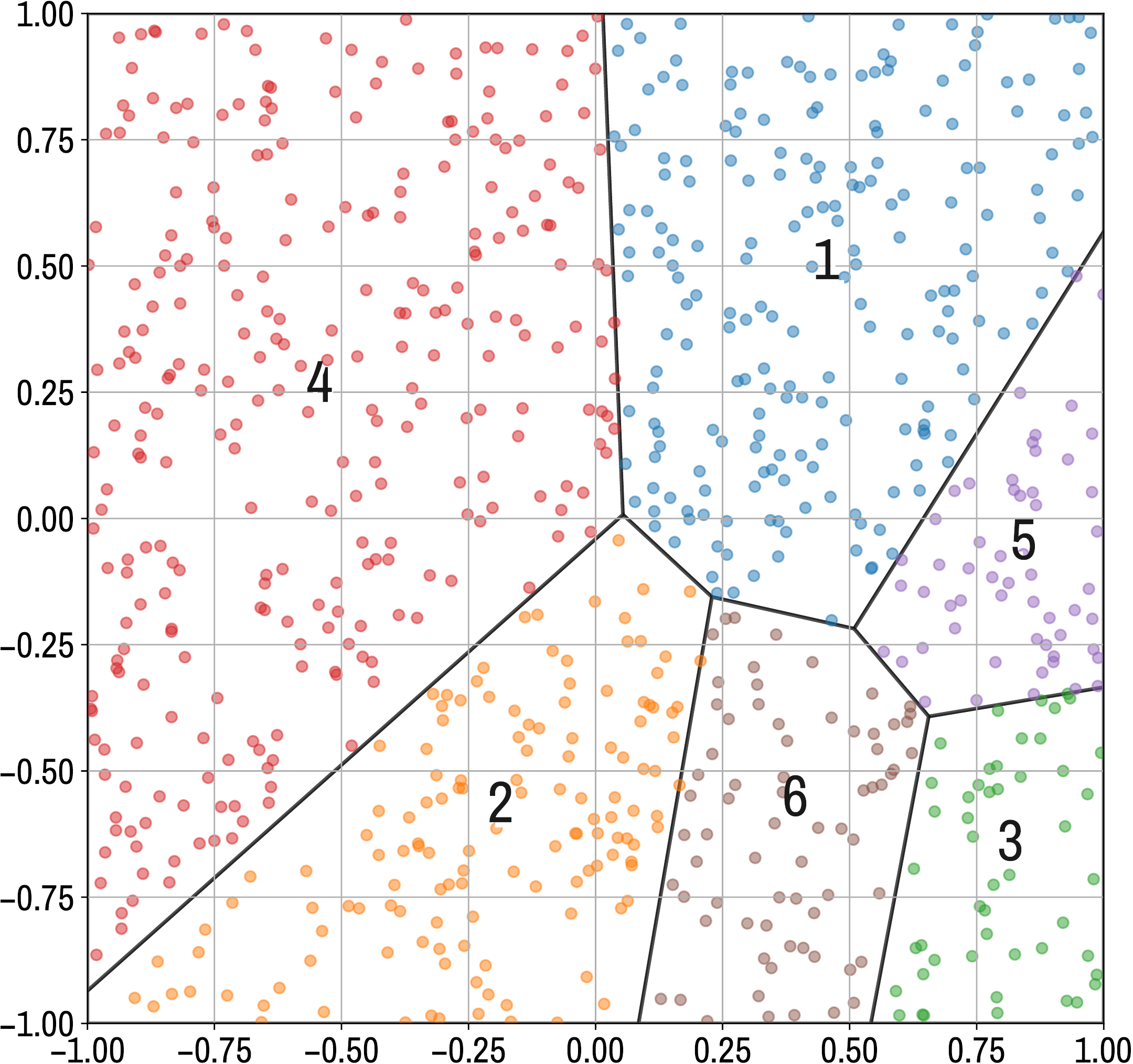}\\~}
                \label{fig:PWA_function_2}}\\
        \end{tabular}
    }
    \label{fig:PWA_example}
    \caption{PARC algorithm for regression on training data generated by the PWA function~\eqref{eq:PWA-example-fun}.}
\end{figure}

We first test whether PARC can reconstruct targets generated from 
the following randomly-generated PWA function
\begin{eqnarray}
    f(x)&=&\max\left\{
    \smallmat{0.8031\\0.0219\\-0.3227}'\smallmat{x_1\\x_2\\1}
    ,
    \smallmat{0.2458\\-0.5823\\-0.1997}'\smallmat{x_1\\x_2\\1}
    ,
    \smallmat{0.0942\\-0.5617\\-0.1622}'\smallmat{x_1\\x_2\\1}
    ,
    \smallmat{0.9462\\-0.7299\\-0.7141}'\smallmat{x_1\\x_2\\1}
    ,
    \right.\nonumber\\&&\left.
    \smallmat{-0.4799\\0.1084\\-0.1210}'\smallmat{x_1\\x_2\\1}
    ,
    \smallmat{0.5770\\0.1574\\-0.1788}'\smallmat{x_1\\x_2\\1}
    \right\}
    \label{eq:PWA-example-fun}
\end{eqnarray}
We generate a dataset of 1000 random samples uniformly distributed
in the box $[-1,1]\times[-1,1]$, plotted in Figures~\ref{fig:PWA_function_3}
and~\ref{fig:PWA_function_1},
from which we extract $N=800$ training
samples and leave the remaining $N=200$ samples for testing.
Figure~\ref{fig:PWA_function_2} shows the partition generated
by the PWL function~\eqref{eq:PWA-example-fun} as in~\eqref{eq:PWL-partition}.

We run PARC with $K=6$, $\sigma=0$,  PWL partitioning~\eqref{eq:softmax}
and $\beta=10^{-3}$, stopping tolerance $\epsilon=10^{-4}$ on 
$V(a,b,\omega,\gamma,z)$, which 
converges in $2.2$~s after 8 iterations. The sequence of function values $V$ 
is 
reported
in Figure~\ref{fig:PWA_function_4}. The final polyhedral partition obtained by PARC
is shown in Figure~\ref{fig:PWA_function_2}. In this ideal case, PARC can 
recover the underlying function generating the data quite well.

\subsubsection{Nonlinear function}
\begin{figure}[t]
    \centerline{
        \begin{tabular}[t]{c} 
            \subfigure[]
            {\parbox{.3\hsize}{\includegraphics[width=\hsize]{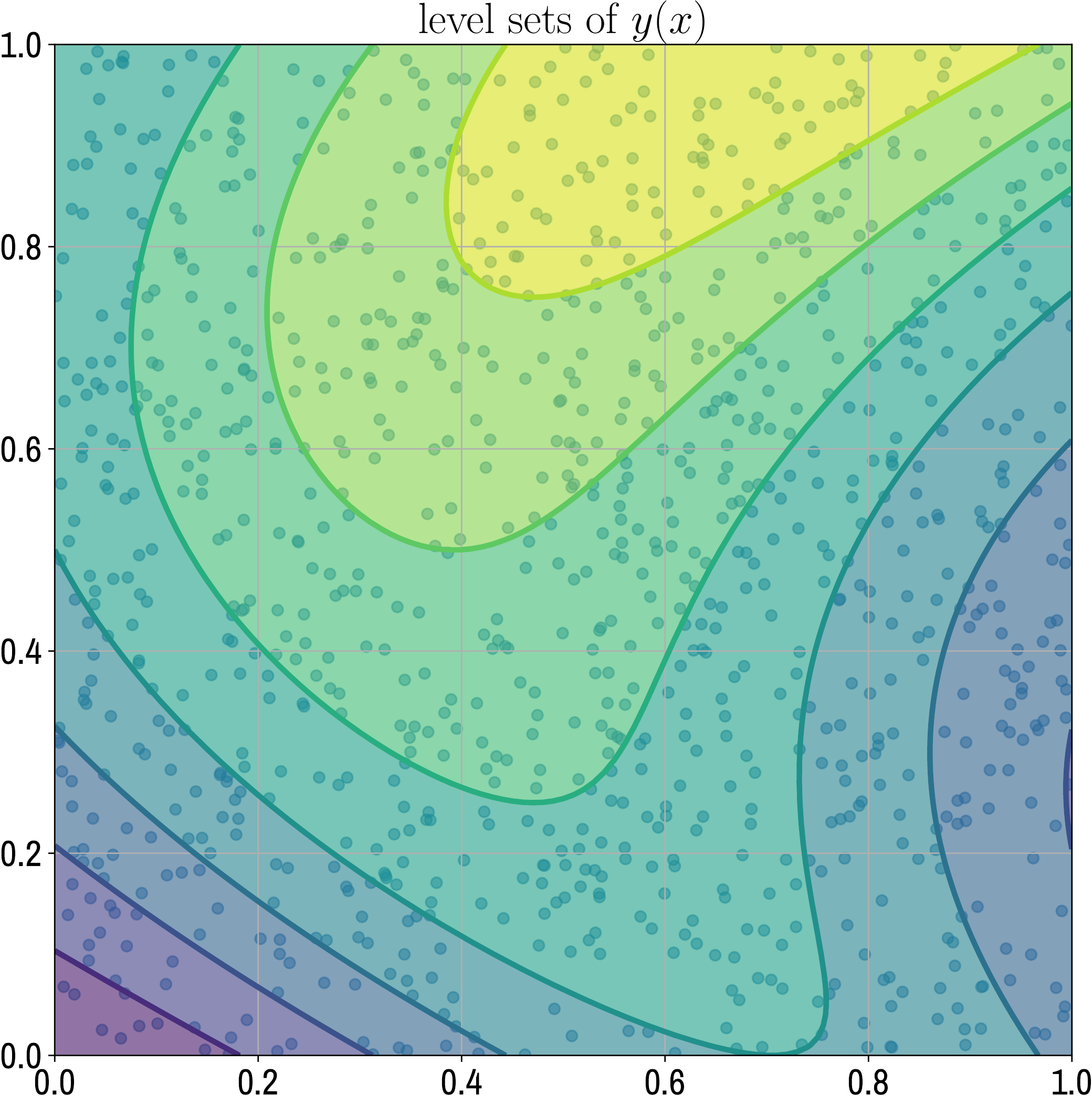}\\~}
                \label{fig:NL_function}}
            \hspace{.05\hsize}
            \subfigure[]
            {\parbox{.3\hsize}{\includegraphics[width=\hsize]{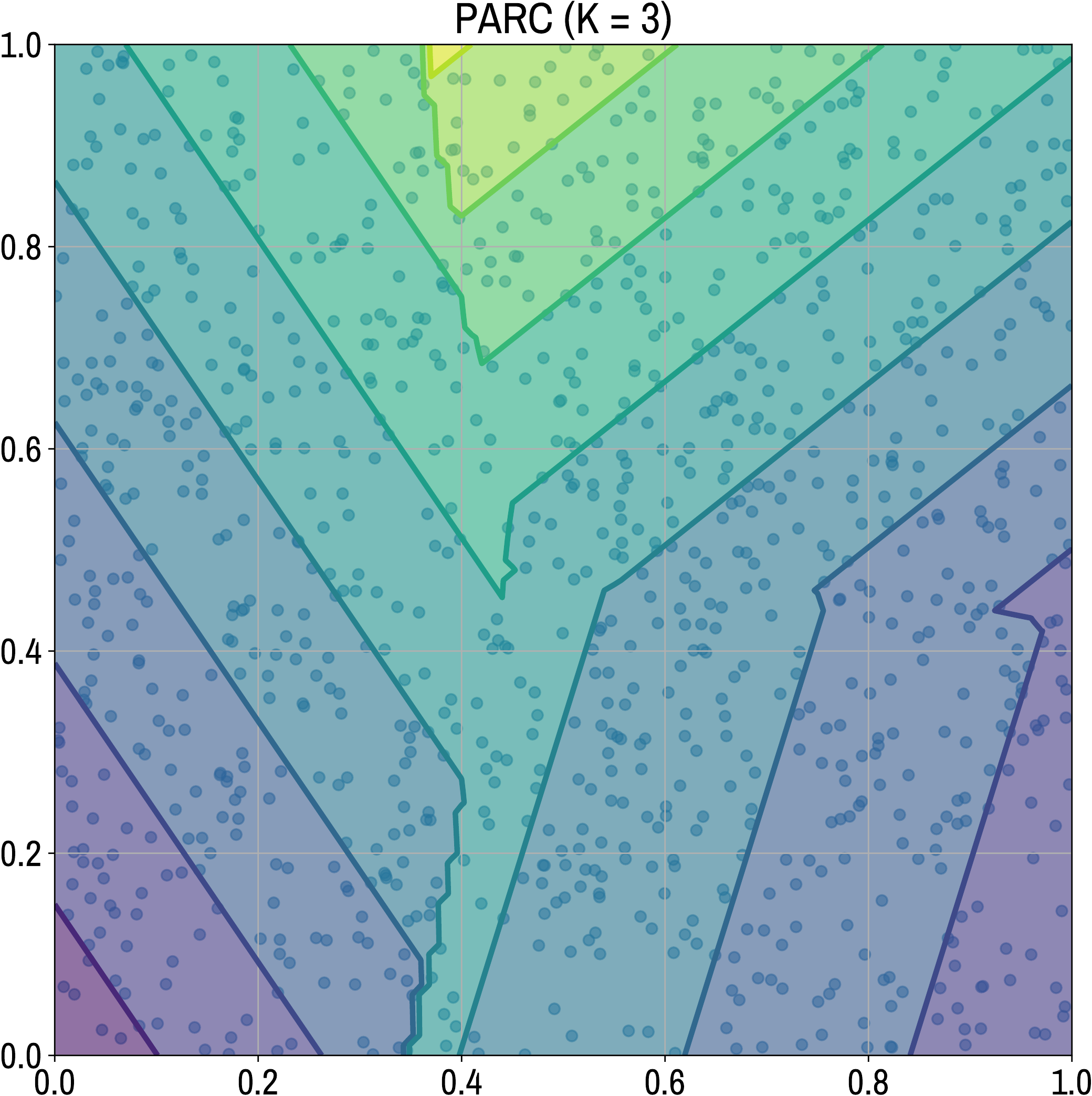}\\~}
                \label{fig:NL_function_3}}
            \hspace{.05\hsize}
            \subfigure[]
            {\parbox{.3\hsize}{\includegraphics[width=\hsize]{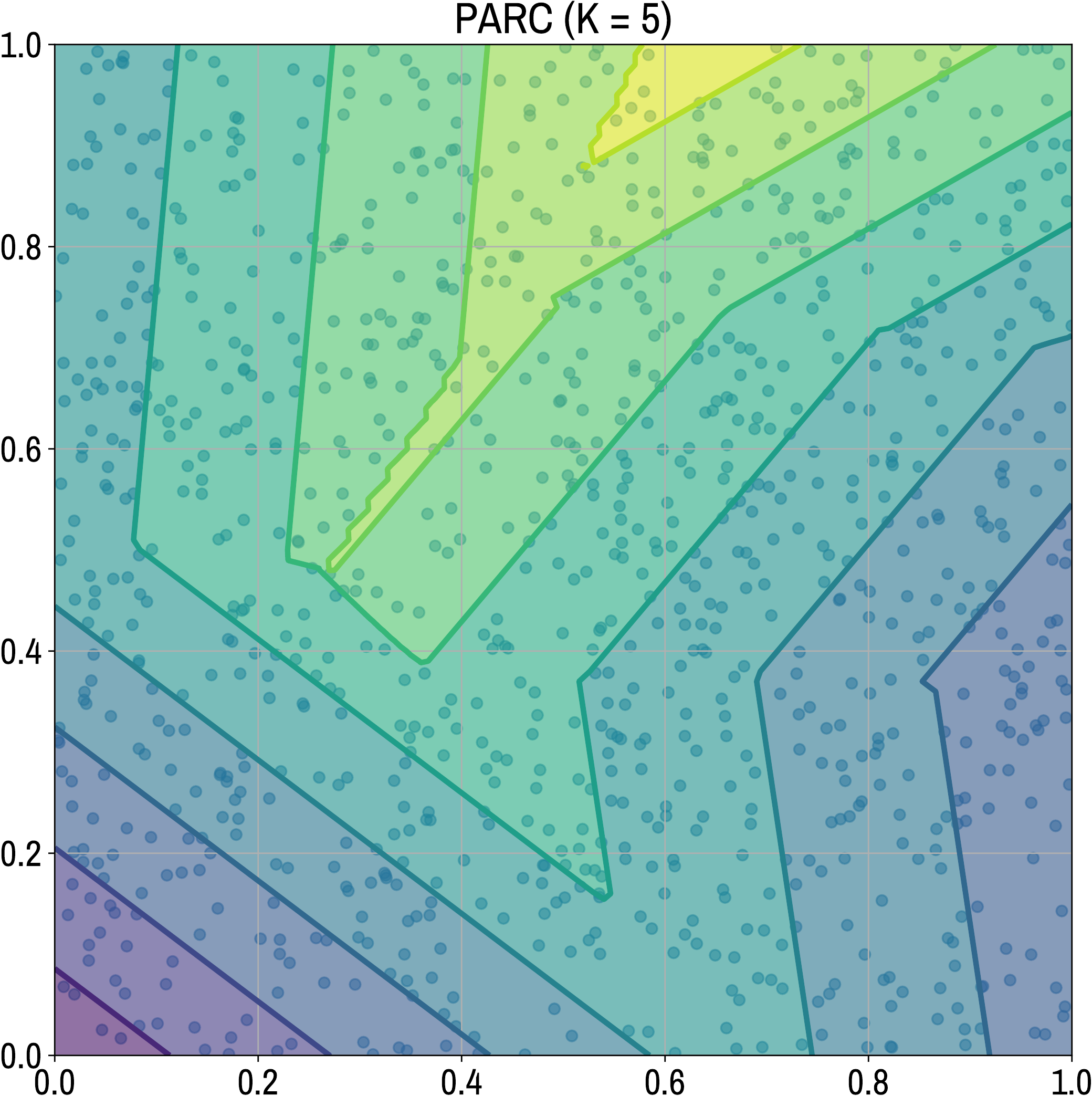}\\~}
                \label{fig:NL_function_5}}
            \\
            \subfigure[]
            {\parbox{.3\hsize}{\includegraphics[width=\hsize]{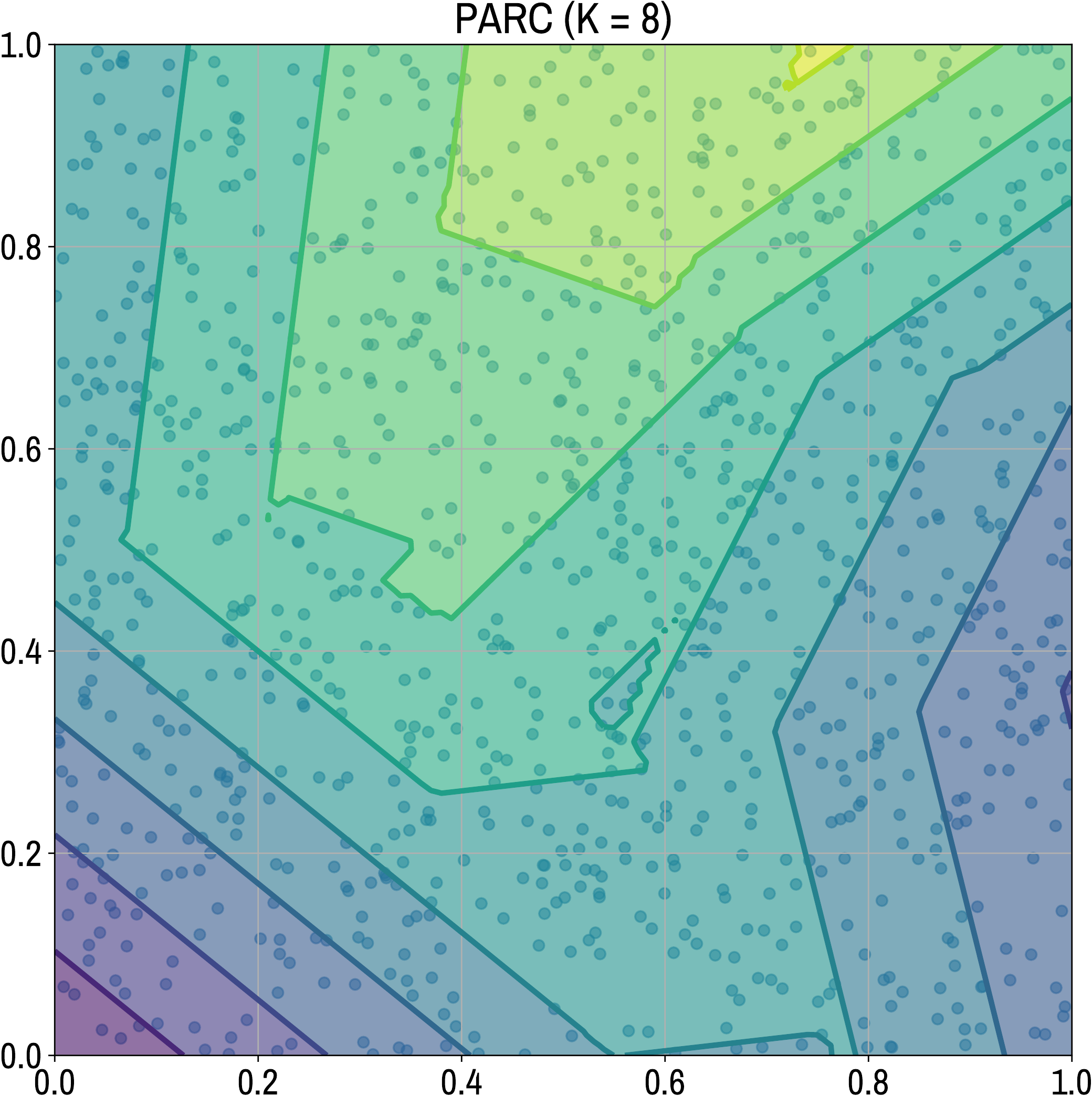}\\~}
            \label{fig:NL_function_8}}
            \hspace{.05\hsize}
            \subfigure[]
            {\parbox{.3\hsize}{\includegraphics[width=\hsize]{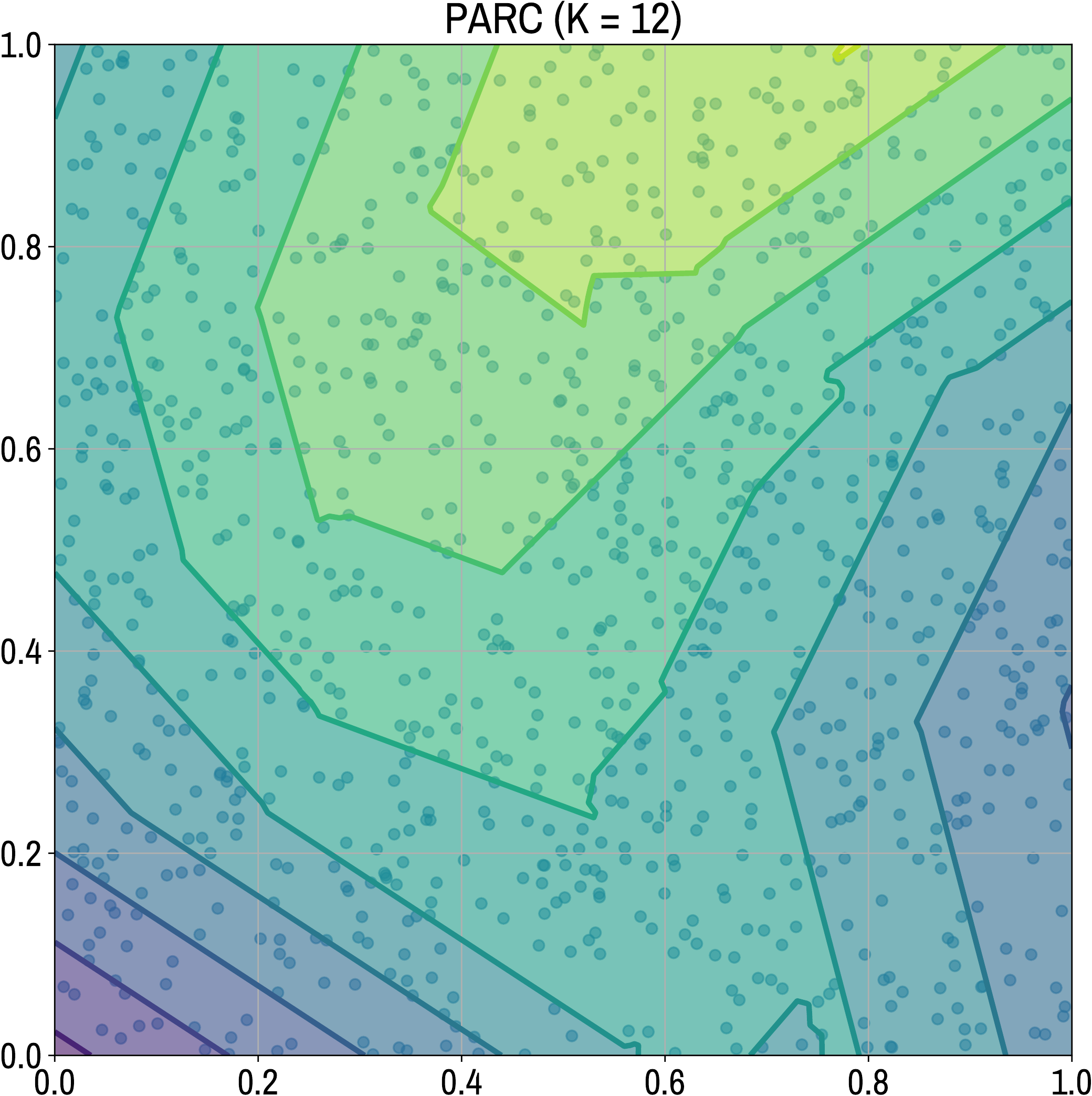}\\~}
            \label{fig:NL_function_12}}
            \hspace{.05\hsize}
            \subfigure[]
            {\parbox{.3\hsize}{\includegraphics[width=\hsize]{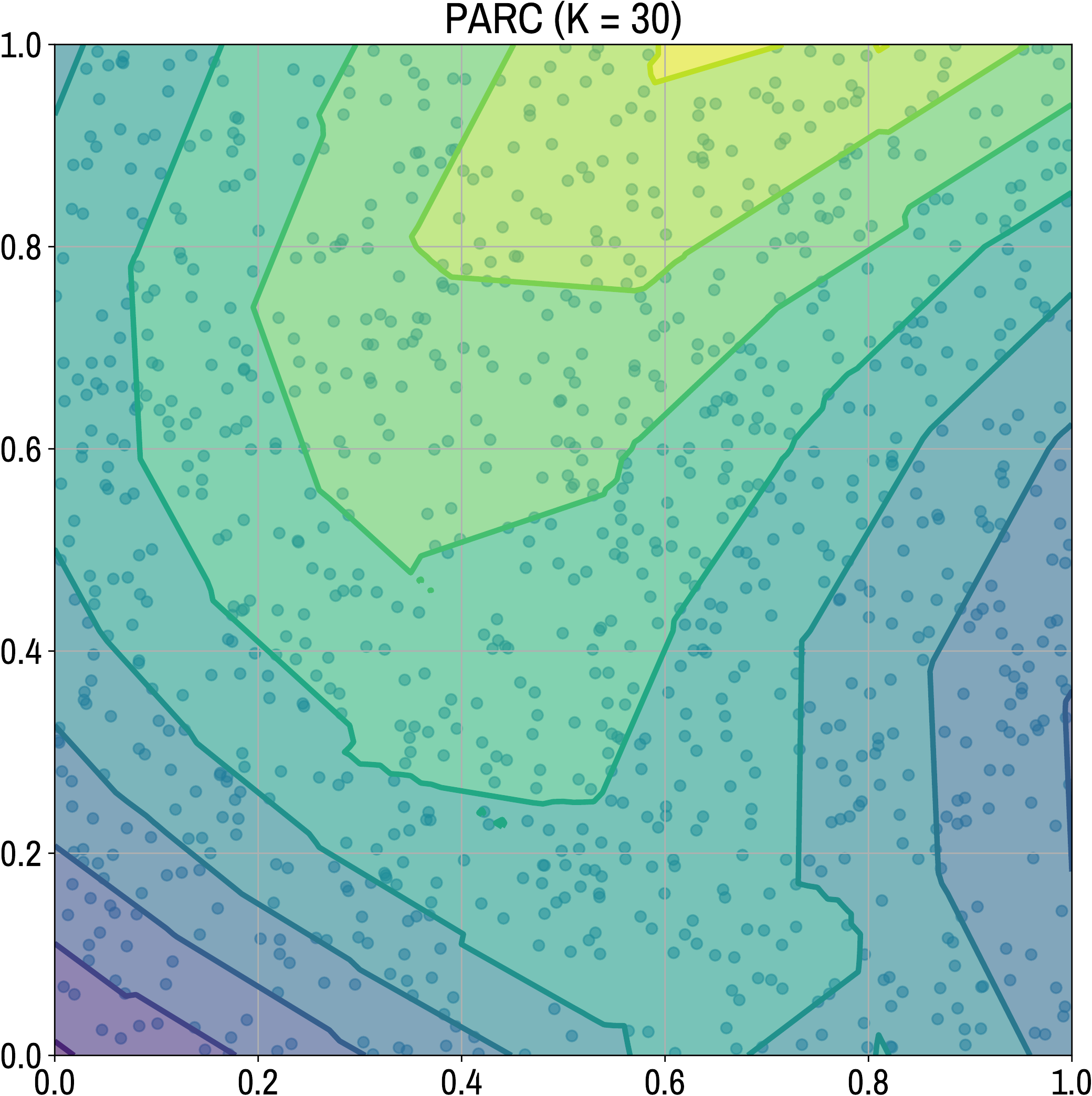}\\~}
            \label{fig:NL_function_30}}
        \end{tabular}
    }
    \caption{Training data and results of PARC for regression: nonlinear 
    function~\eqref{eq:NL-example-function}.}
    \label{fig:NL_example-1}    
\end{figure}

\begin{figure}[th!]
    \centerline{
        \begin{tabular}[t]{c} 
            \subfigure[]
            {\parbox{.3\hsize}{\includegraphics[width=\hsize]{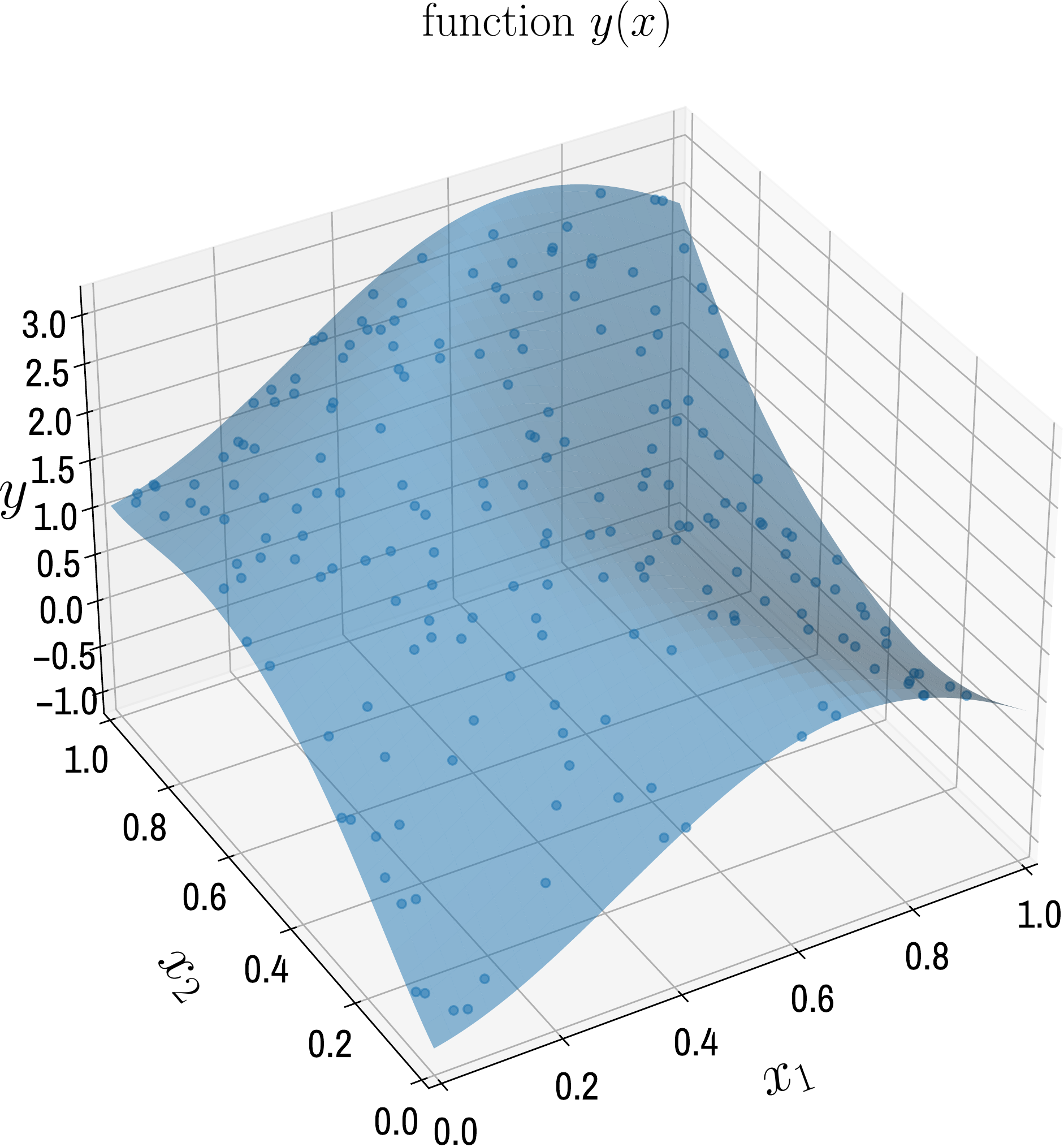}\\~}
                \label{fig:NL_function_3d}}
            \hspace{.05\hsize}
            \subfigure[]
            {\parbox{.3\hsize}{\includegraphics[width=\hsize]{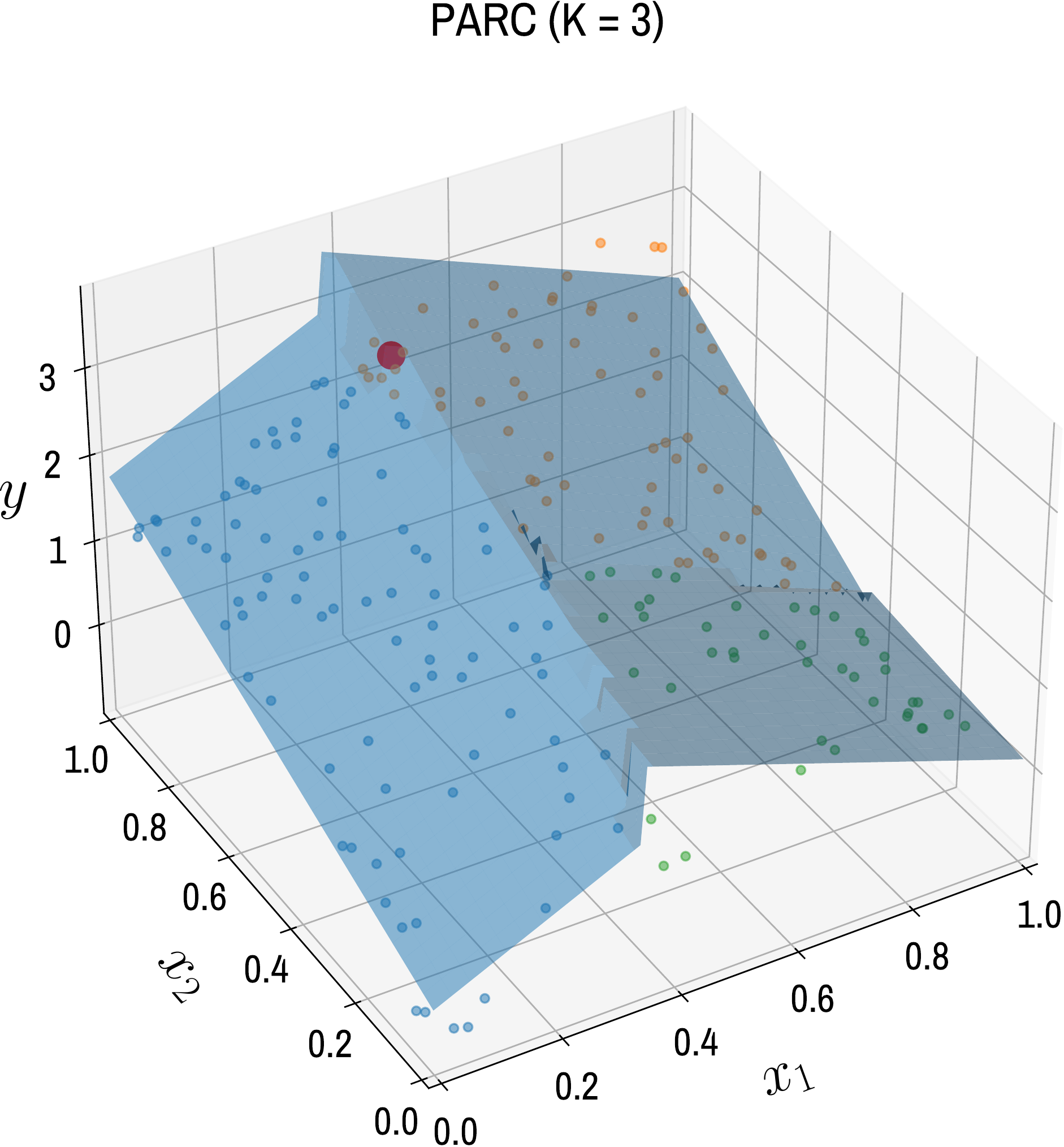}\\~}
                \label{fig:NL_function_3_3d}}
            \hspace{.05\hsize}
            \subfigure[]
            {\parbox{.3\hsize}{\includegraphics[width=\hsize]{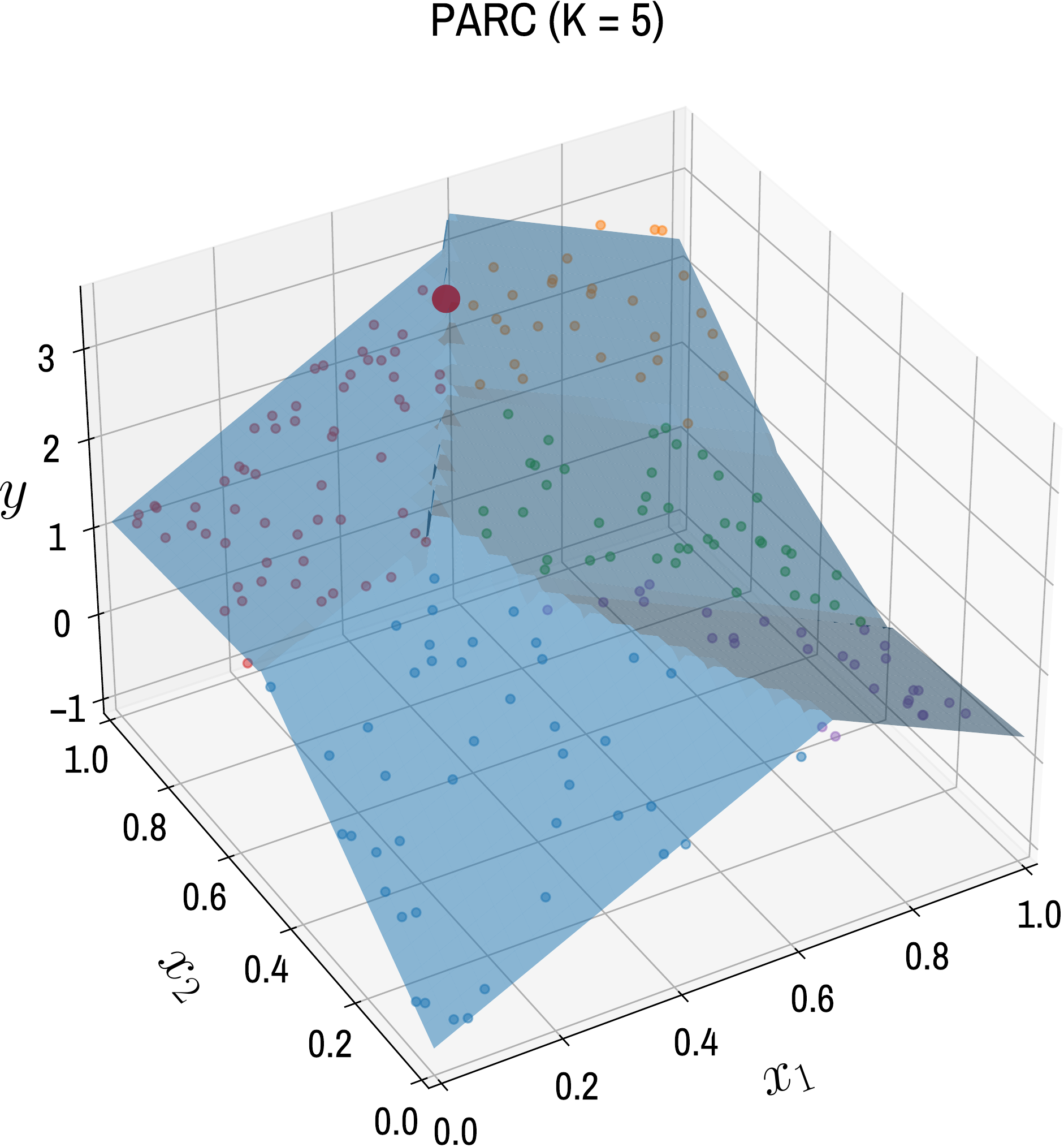}\\~}
                \label{fig:NL_function_5_3d}}
            \\
            \subfigure[]
            {\parbox{.3\hsize}{\includegraphics[width=\hsize]{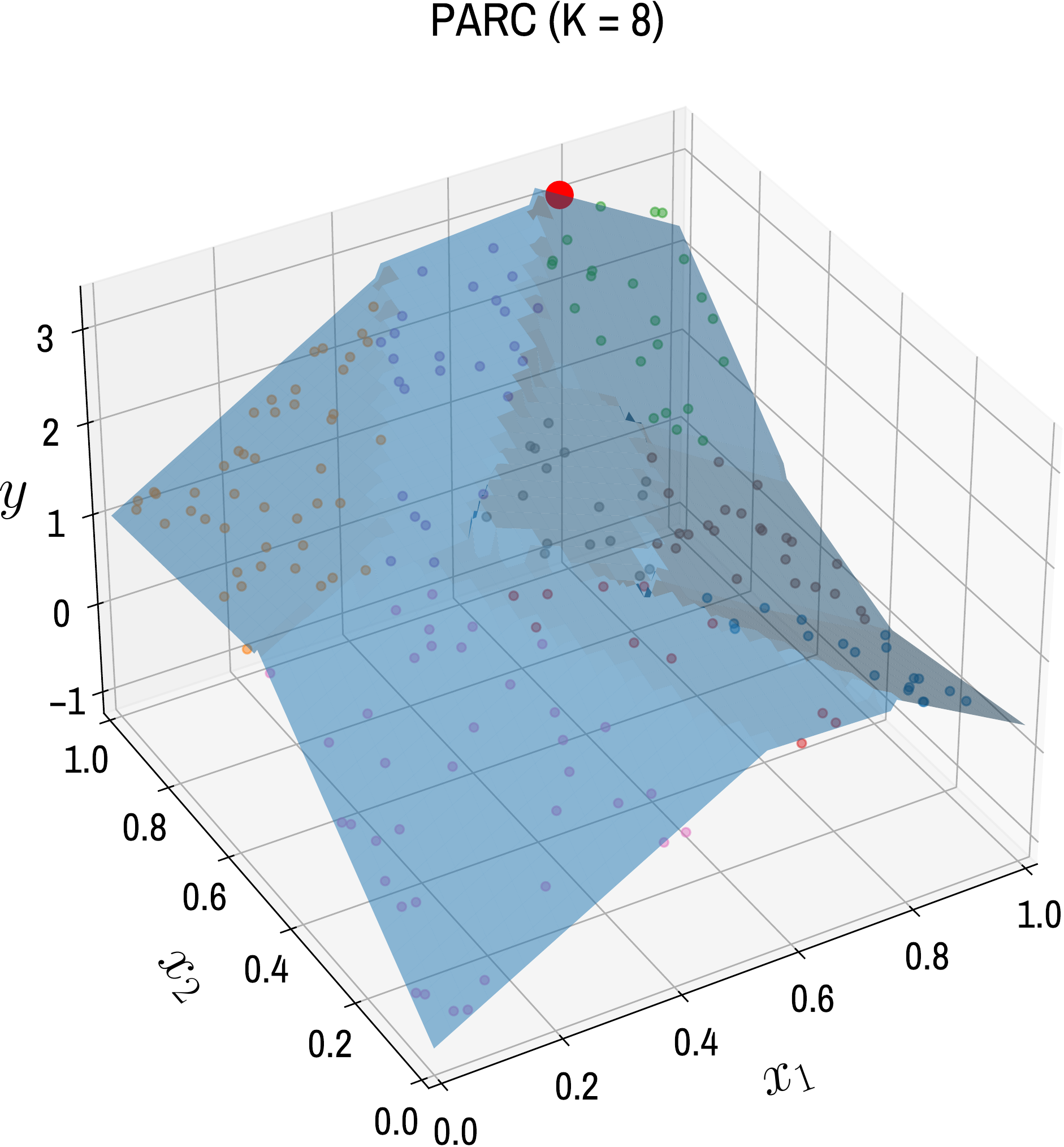}\\~}
                \label{fig:NL_function_8_3d}}
            \hspace{.05\hsize}
            \subfigure[]
            {\parbox{.3\hsize}{\includegraphics[width=\hsize]{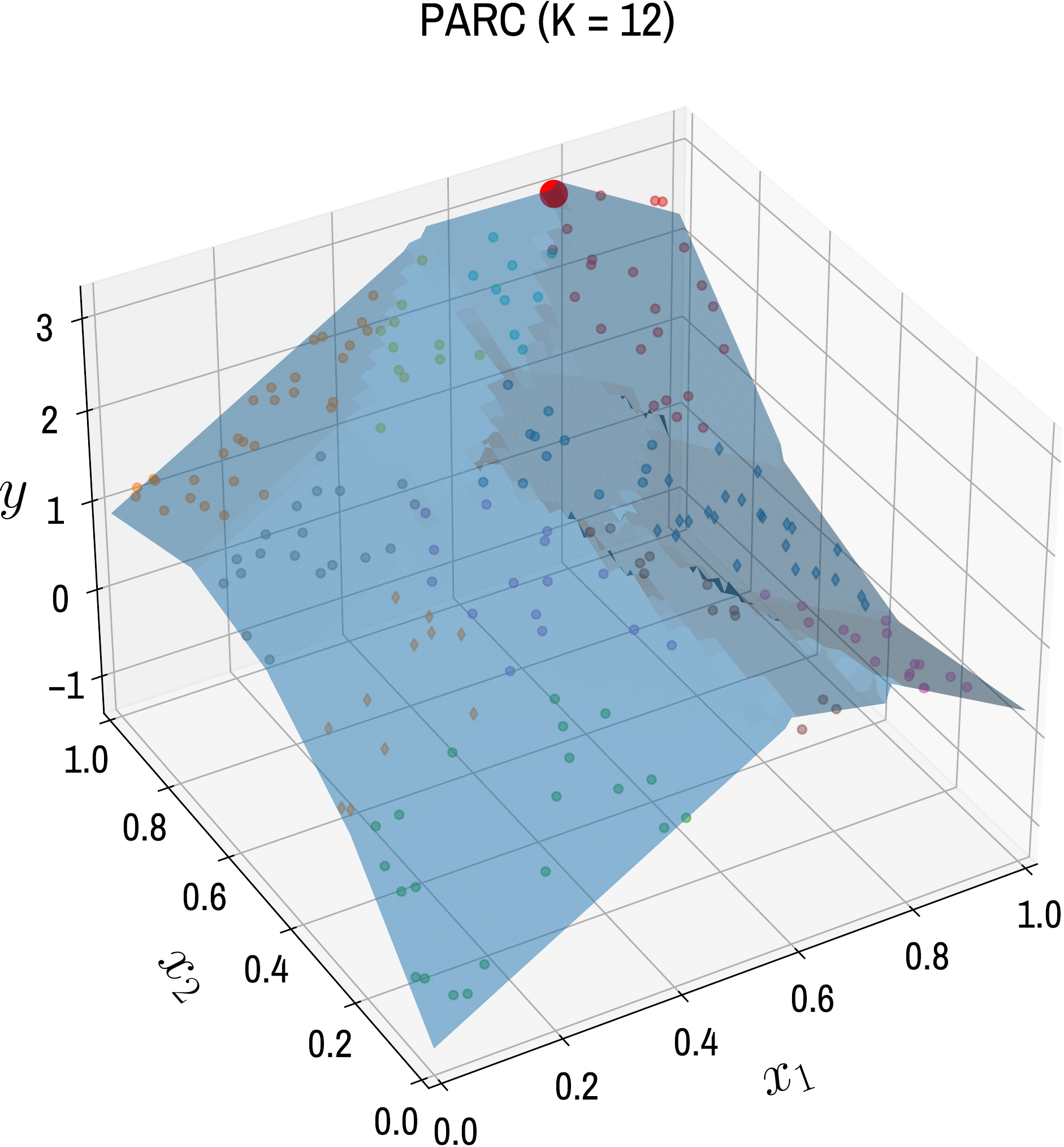}\\~}
                \label{fig:NL_function_12_3d}}
            \hspace{.05\hsize}
            \subfigure[]
            {\parbox{.3\hsize}{\includegraphics[width=\hsize]{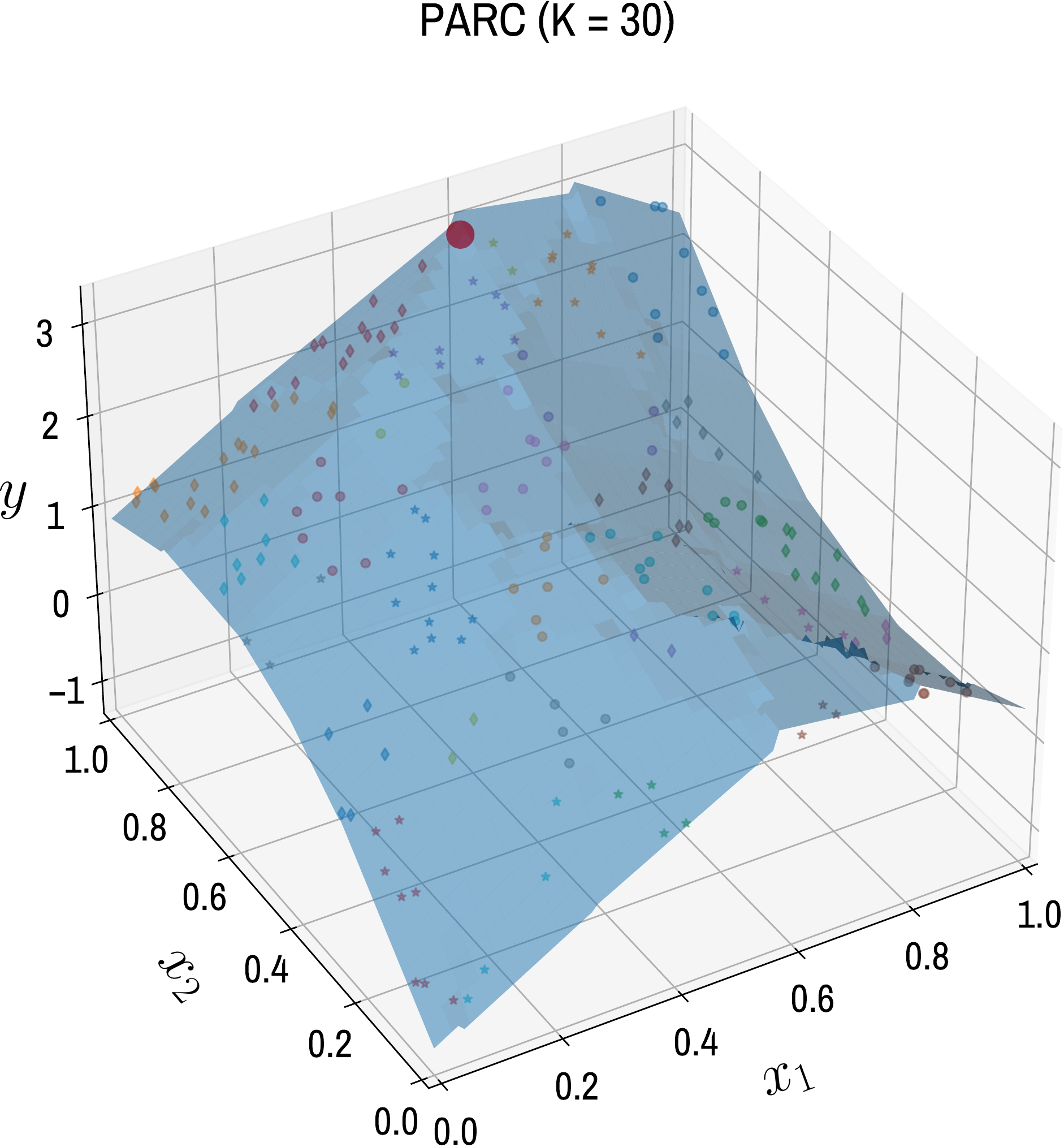}\\~}
                \label{fig:NL_function_30_3d}}
        \end{tabular}
    }
    \caption{Training data and results of PARC for regression: nonlinear 
    function~\eqref{eq:NL-example-function}. The result of the MILP 
    optimization~\eqref{eq:MILP-tracking} is represented by the red dot.}
    \label{fig:NL_example-2}
\end{figure}

We solve another simple regression example on a dataset of $N=1000$ randomly-generated
samples of the nonlinear function 
\begin{equation}
    y(x_1,x_2) = \sin\left(4x_1-5\left(x_2 - \frac{1}{2}\right)^2\right) + 2x_2
    \label{eq:NL-example-function}
\end{equation}
Again we use 80\% of the samples as training data and the remaining 20\% for 
testing. The function and the training dataset are shown in 
Figures~\ref{fig:NL_function},~\ref{fig:NL_function_3d}.
We run PARC with $\sigma=1$, $\epsilon=10^{-4}$, PWL 
partitioning~\eqref{eq:softmax} with $\beta=10^{-3}$, and different values of 
$K$. The level sets and training data are reported in 
Figure~\ref{fig:NL_example-1}.
The resulting piecewise linear regression functions are shown in Figure~\ref{fig:NL_example-2},
which also shows the solution obtained by solving the MILP~\eqref{eq:MILP-tracking}
for $y_{\rm ref}=3$.

The results obtained by running PARC for different values of $K$, $\sigma$ and the
two alternative separation criteria (Voronoi partitioning and softmax 
regression with $\beta=10^{-3}$)
are reported in Table~\ref{tab:NL_R2train} (R$^2$-score on training data),
Table~\ref{tab:NL_R2test} (R$^2$-score on test data),
Table~\ref{tab:NL_CPUtrain} (CPU time [s] to execute PARC),
Table~\ref{tab:NL_Iters} (number of PARC iterations). The best results are
usually obtained for $\sigma=1$ using softmax regression (S) for PWL partitioning
as in~\eqref{eq:softmax-cost}.

The CPU time spent to solve the MILP~\eqref{eq:MILP-tracking} using the CBC 
solver\footnote{\url{https://github.com/coin-or/Cbc}}
through the Python MIP 
package~\footnote{\url{https://github.com/coin-or/python-mip}} 
for $K$ = 3, 5, 8, 12, and 30 is, respectively, 8, 29, 85, 251, and
1420~ms.
Note that the case $K=1$ corresponds to ridge regression on the entire dataset, 
while $\sigma=10000$ approximates the case $\sigma\rightarrow +\infty$, corresponding
to pure PWL separation + ridge regression on each cluster.

\begin{table}[t]\begin{center}
        \scalebox{.8}[0.8]{
        \begin{tabular}{r|r|r|r|r|r|r}
            $\sigma$ & $K=1$~~~~ & $K=3$~~~~ & $K=5$~~~~ & $K=8$~~~~ & 
            $K=12$~~~~ & $K=30$~~~~ \\[.5em]\hline
            (S)
            \hfill 0
            & 0.565 (1.4\%)& 0.899 (2.0\%)& 0.979 (0.2\%)& 0.991 (0.2\%)& 0.995 
            (0.2\%)& 0.998 (0.1\%) \\
            (V)
            \hfill 0
            & 0.565 (1.4\%)& 0.886 (3.1\%)& 0.974 (0.3\%)& 0.986 (0.2\%)& 0.993 
            (0.2\%)& 0.998 (0.0\%) \\
            (S)
            \hfill 0.01
            & 0.565 (1.4\%)& 0.899 (2.2\%)& 0.979 (0.2\%)& 0.991 (0.2\%)& 0.995 
            (0.1\%)& 0.999 (0.1\%) \\
            (V)
            \hfill 0.01
            & 0.565 (1.4\%)& 0.887 (3.1\%)& 0.973 (0.3\%)& 0.986 (0.2\%)& 0.993 
            (0.1\%)& 0.998 (0.0\%) \\
            (S)
            \hfill 1
            & 0.565 (1.4\%)& 0.895 (2.3\%)& 0.982 (0.2\%)& 0.994 (0.2\%)& 0.998 
            (0.0\%)& 0.999 (0.0\%) \\
            (V)
            \hfill 1
            & 0.565 (1.4\%)& 0.881 (3.0\%)& 0.974 (0.3\%)& 0.986 (0.2\%)& 0.994 
            (0.1\%)& 0.999 (0.0\%) \\
            (S)
            \hfill 100
            & 0.565 (1.4\%)& 0.908 (0.9\%)& 0.977 (0.5\%)& 0.986 (0.2\%)& 0.994 
            (0.1\%)& 0.999 (0.0\%) \\
            (V)
            \hfill 100
            & 0.565 (1.4\%)& 0.887 (3.6\%)& 0.972 (0.4\%)& 0.989 (0.3\%)& 0.995 
            (0.0\%)& 0.999 (0.0\%) \\
            (S)
            \hfill 10000
            & 0.565 (1.4\%)& 0.834 (2.1\%)& 0.969 (0.3\%)& 0.985 (0.2\%)& 0.994 
            (0.1\%)& 0.999 (0.0\%) \\
            (V)
            \hfill 10000
            & 0.565 (1.4\%)& 0.865 (3.6\%)& 0.971 (0.3\%)& 0.985 (0.2\%)& 0.994 
            (0.1\%)& 0.999 (0.0\%) \\
            \hline
        \end{tabular}}
        \caption{PARC regression on targets from nonlinear 
            function~\eqref{eq:NL-example-function}: R$^2$ score on training 
            data, mean (std).
            PWL separation: (S) = softmax regression, (V) for Voronoi 
            partitioning.}
        \label{tab:NL_R2train}
    \end{center}
\end{table}
\begin{table}[t]\begin{center}
        \scalebox{.8}[0.8]{
        \begin{tabular}{r|r|r|r|r|r|r}
            $\sigma$ & $K=1$~~~~ & $K=3$~~~~ & $K=5$~~~~ & $K=8$~~~~ & 
            $K=12$~~~~ & $K=30$~~~~ \\[.5em]\hline
            (S)
            \hfill 0
            & 0.548 (6.5\%)& 0.889 (2.6\%)& 0.976 (0.5\%)& 0.989 (0.3\%)& 0.994 
            (0.2\%)& 0.997 (0.1\%) \\
            (V)
            \hfill 0
            & 0.548 (6.5\%)& 0.872 (3.6\%)& 0.970 (0.7\%)& 0.985 (0.4\%)& 0.993 
            (0.2\%)& 0.998 (0.1\%) \\
            (S)
            \hfill 0.01
            & 0.548 (6.5\%)& 0.894 (2.5\%)& 0.976 (0.5\%)& 0.989 (0.4\%)& 0.994 
            (0.1\%)& 0.998 (0.1\%) \\
            (V)
            \hfill 0.01
            & 0.548 (6.5\%)& 0.877 (3.3\%)& 0.969 (0.6\%)& 0.985 (0.3\%)& 0.992 
            (0.3\%)& 0.997 (0.1\%) \\
            (S)
            \hfill 1
            & 0.548 (6.5\%)& 0.883 (2.8\%)& 0.981 (0.3\%)& 0.993 (0.2\%)& 0.997 
            (0.1\%)& 0.999 (0.0\%) \\
            (V)
            \hfill 1
            & 0.548 (6.5\%)& 0.868 (3.5\%)& 0.970 (0.7\%)& 0.985 (0.3\%)& 0.993 
            (0.2\%)& 0.998 (0.1\%) \\
            (S)
            \hfill 100
            & 0.548 (6.5\%)& 0.898 (1.6\%)& 0.970 (1.0\%)& 0.982 (0.2\%)& 0.992 
            (0.2\%)& 0.998 (0.0\%) \\
            (V)
            \hfill 100
            & 0.548 (6.5\%)& 0.874 (4.1\%)& 0.967 (0.8\%)& 0.987 (0.4\%)& 0.993 
            (0.2\%)& 0.998 (0.1\%) \\
            (S)
            \hfill 10000
            & 0.548 (6.5\%)& 0.816 (3.2\%)& 0.963 (0.8\%)& 0.980 (0.3\%)& 0.992 
            (0.1\%)& 0.998 (0.0\%) \\
            (V)
            \hfill 10000
            & 0.548 (6.5\%)& 0.846 (4.4\%)& 0.965 (0.8\%)& 0.982 (0.3\%)& 0.993 
            (0.2\%)& 0.998 (0.0\%) \\
            \hline
        \end{tabular}}
        \caption{PARC regression on targets from nonlinear 
            function~\eqref{eq:NL-example-function}: R$^2$ score on test data, 
            mean (std).
            PWL separation: (S) = softmax regression, (V) for Voronoi 
            partitioning.}
        \label{tab:NL_R2test}
    \end{center}
\end{table}

\begin{table}[t]\begin{center}
        \scalebox{.8}[0.8]{
        \begin{tabular}{r|r|r|r|r|r|r}
            $\sigma$ & $K=1$~~~~ & $K=3$~~~~ & $K=5$~~~~ & $K=8$~~~~ & 
            $K=12$~~~~ & $K=30$~~~~ \\[.5em]\hline
            (S)
            \hfill 0
            &   1.0 (0.0\%)&  18.9 (44.6\%)&  13.1 (27.3\%)&  16.9 (28.4\%)&  
            18.9 (22.8\%)&  13.0 (17.9\%) \\
            (V)
            \hfill 0
            &   1.0 (0.0\%)&  20.2 (39.7\%)&  12.8 (26.3\%)&  17.1 (30.4\%)&  
            20.1 (27.4\%)&  13.9 (20.5\%) \\
            (S)
            \hfill 0.01
            &   1.0 (0.0\%)&  17.7 (42.4\%)&  13.3 (37.0\%)&  17.4 (28.3\%)&  
            20.5 (39.6\%)&  12.3 (19.6\%) \\
            (V)
            \hfill 0.01
            &   1.0 (0.0\%)&  17.8 (43.0\%)&  13.8 (32.7\%)&  14.3 (26.3\%)&  
            19.7 (31.5\%)&  14.5 (33.8\%) \\
            (S)
            \hfill 1
            &   1.0 (0.0\%)&  19.2 (46.0\%)&  11.2 (27.2\%)&  15.5 (27.3\%)&  
            14.2 (17.8\%)&   7.9 (14.9\%) \\
            (V)
            \hfill 1
            &   1.0 (0.0\%)&  19.4 (41.4\%)&  13.0 (39.1\%)&  15.1 (23.1\%)&  
            18.9 (36.4\%)&  12.5 (33.2\%) \\
            (S)
            \hfill 100
            &   1.0 (0.0\%)&  19.4 (24.1\%)&   8.2 (36.3\%)&   5.8 (32.5\%)&   
            4.0 (37.8\%)&   5.2 (24.0\%) \\
            (V)
            \hfill 100
            &   1.0 (0.0\%)&  17.4 (49.1\%)&  11.4 (41.7\%)&  17.2 (42.6\%)&  
            12.8 (23.5\%)&   8.9 (22.7\%) \\
            (S)
            \hfill 10000
            &   1.0 (0.0\%)&   3.0 (31.2\%)&   3.1 (26.8\%)&   3.4 (30.0\%)&   
            4.6 (27.0\%)&   5.5 (29.9\%) \\
            (V)
            \hfill 10000
            &   1.0 (0.0\%)&  11.7 (53.7\%)&   5.9 (45.8\%)&   4.5 (29.7\%)&   
            4.2 (37.1\%)&   3.5 (64.5\%) \\
            \hline
        \end{tabular}}
        \caption{PARC regression on targets from nonlinear 
            function~\eqref{eq:NL-example-function}: number of PARC iterations, 
            mean (std).
            PWL separation: (S) = softmax regression, (V) for Voronoi 
            partitioning.}
        \label{tab:NL_Iters}
    \end{center}
\end{table}

\begin{table}[t]\begin{center}
        \scalebox{.8}[0.8]{
        \begin{tabular}{r|r|r|r|r|r|r}
            $\sigma$ & $K=1$~~~~ & $K=3$~~~~ & $K=5$~~~~ & $K=8$~~~~ & 
            $K=12$~~~~ & $K=30$~~~~ \\[.5em]\hline
            (S)
            \hfill 0
            &  0.12 (9.8\%)&  1.33 (43.6\%)&  1.46 (24.6\%)&  2.90 (28.0\%)&  
            5.18 (21.1\%)&  7.14 (17.3\%) \\
            (V)
            \hfill 0
            &  0.04 (11.4\%)&  0.78 (36.1\%)&  0.77 (24.9\%)&  1.58 (30.7\%)&  
            2.64 (26.3\%)&  4.44 (19.5\%) \\
            (S)
            \hfill 0.01
            &  0.12 (7.7\%)&  1.25 (40.3\%)&  1.48 (36.4\%)&  2.95 (30.1\%)&  
            5.47 (40.9\%)&  7.22 (18.7\%) \\
            (V)
            \hfill 0.01
            &  0.04 (13.2\%)&  0.65 (41.2\%)&  0.82 (31.4\%)&  1.31 (25.1\%)&  
            2.58 (30.8\%)&  4.61 (32.8\%) \\
            (S)
            \hfill 1
            &  0.12 (10.4\%)&  1.36 (43.8\%)&  1.42 (24.7\%)&  3.36 (26.0\%)&  
            4.14 (15.9\%)&  4.26 (14.9\%) \\
            (V)
            \hfill 1
            &  0.04 (13.5\%)&  0.71 (38.6\%)&  0.78 (35.7\%)&  1.36 (22.7\%)&  
            2.48 (36.0\%)&  3.98 (31.1\%) \\
            (S)
            \hfill 100
            &  0.12 (8.9\%)&  1.45 (24.2\%)&  1.11 (34.5\%)&  1.10 (30.4\%)&  
            1.04 (35.6\%)&  2.68 (22.1\%) \\
            (V)
            \hfill 100
            &  0.04 (10.6\%)&  0.64 (44.9\%)&  0.69 (38.0\%)&  1.56 (43.5\%)&  
            1.70 (21.9\%)&  2.90 (21.0\%) \\
            (S)
            \hfill 10000
            &  0.12 (9.2\%)&  0.24 (27.1\%)&  0.41 (22.6\%)&  0.66 (26.9\%)&  
            1.14 (24.2\%)&  2.80 (27.2\%) \\
            (V)
            \hfill 10000
            &  0.04 (11.5\%)&  0.45 (49.7\%)&  0.38 (39.7\%)&  0.45 (24.8\%)&  
            0.62 (31.4\%)&  1.21 (61.8\%) \\
            \hline
        \end{tabular}}
        \caption{PARC regression on targets from nonlinear 
            function~\eqref{eq:NL-example-function}: training time [s], mean 
            (std).
            PWL separation: (S) = softmax regression, (V) for Voronoi 
            partitioning.}
        \label{tab:NL_CPUtrain}
    \end{center}
\end{table}

\subsection{Real-world datasets}
\label{sec:real-datasets}
We test the PARC algorithm on real-world datasets for regression
and classification from the PMLB repository~\citep{Olson2017PMLB}.
The features containing four or less distinct values are treated
as categorical and one-hot encoded, all the remaining features as numerical.
In all tests, $N_{\rm tot}$ denotes the total number of samples in the
dataset, whose 80\% is used for training and the rest 20\% for testing.
PARC is run with $\sigma=1$, softmax regression for PWL partitioning~\eqref{eq:softmax-cost},
$\epsilon=10^{-4}$, $\alpha=0.1$, $\beta=10^{-3}$. The minimum size of a 
cluster not 
to be discarded is 1\% of the number $N$ of training samples. Prediction quality
is measured in terms of $R^2$ score (in case of regression 
problems), or accuracy score $a$ (for classification), respectively defined 
as
\[
    R^2=1-\frac{\sum_{k=1}^N(y_k-\hat y(x_k))^2}{\sum_{k=1}^N(y_k-\frac{1}{N}\sum_{k=1}^Ny_k)^2},\quad
    a=\frac{1}{N}\sum_{k=1}^N [\hat y(x_k)=y_k]
\]
The neural networks and decision trees used for comparison are trained using
scikit-learn~\citep{scikit-learn} functions. The stochastic optimizer 
Adam~\citep{KB15} is used for training the coefficient and bias terms of the 
neural 
network.

\subsubsection{Regression problems}
We extracted all the datasets from the PMLB repository with numeric targets 
containing between $N_{\rm 
tot}=500$ and $5000$ samples and between $n_x=2$ and $20$ features (before 
one-hot 
encoding categorical features). Five-fold cross-validation is run on 
the training dataset for all values of $K$ between $2$ and $15$ to determine 
the value $K^*$ that is optimal in terms of average $R^2$ score. For 
comparison, we run PARC with fixed values of $K$ and compare against other 
methods providing 
piecewise linear partitions, particularly a neural network with ReLU 
activation function with a single layer of $K^*$ neurons and a decision tree 
with ten non-leaf nodes. Note that the neural network requires $K^*$ 
binary variables to encode the ReLU activation functions in a MIP, the same 
number as the PWL 
regressor determined by PARC, 
as described in Section~\ref{sec:MIP}. In contrast, the decision tree
requires ten binary variables.

The $R^2$ scores obtained on the datasets are shown in 
Table~\ref{tab:dataset-regression-training} 
(training data) and in Table~\ref{tab:dataset-regression-test} (test data).
The CPU time spent on solving the training problems is reported in 
Table~\ref{tab:dataset-CPUTIME-regression}.

The results show that PARC often provides better fit on training data, 
especially for large values of $K$. On test data, PARC and neural networks 
with $K^*$ ReLU neurons provide the best results. Some poor results of PARC on
test data for large values of $K$ are usually associated with overfitting 
the training dataset, see for example \textsf{522\_pm10}, \textsf{547\_no2}, 
\textsf{627\_fri\_C1\_1000\_5}.

\subsubsection{Classification problems}
We extracted all datasets from the PMLB repository with categorical targets 
with at most $m_1=10$ 
classes, containing between $N_{\rm tot}=1000$ and 5000 samples, and between 
$n_x=2$ and 20 features (before one-hot encoding categorical features). 
We compare PARC with $K=2$, $3$, $5$ to softmax regression (corresponding to 
setting $K=1$ in PARC), a neural network (NN) with ReLU activation function and 
a single layer of $K=5$ neurons, and a decision tree (DT) with $5$ non-leaf 
nodes. Encoding the PARC classifier as an MIP requires $K+m_1$ binary variables 
as described in Section~\ref{sec:MIP}, the NN requires $5+m_1$ binary variables 
for MIP encoding, the DT requires $5+m_1$ binary variables ($m_1$ 
variables are required to encode the $\arg\max$ selecting the class
with highest score). In this test campaign, computing $K^*$ by cross-validation 
has not shown to bring significant benefits and is not reported.

The accuracy scores obtained on the datasets are shown in 
Table~\ref{tab:dataset-classification-training} 
(training data) and in Table~\ref{tab:dataset-classification-test} (test data).
The CPU time spent on solving the training problems is reported in 
Table~\ref{tab:dataset-CPUTIME-classification}. On training data, PARC with 
$K=5$ provides the best accuracy in about 75\% 
of the datasets, with neural networks based on $5$ ReLU neurons better 
performing in the remaining cases. On test data, most methods perform 
similarly, with neural networks providing slightly superior accuracy.

\section{Conclusions}
\label{sec:conclusions}
The regression and classification algorithm proposed in this paper generalizes
linear regression and classification approaches, in particular, ridge 
regression and softmax 
regression, to a piecewise linear form. Such a form is amenable for 
mixed-integer encoding, particularly beneficial when the obtained 
predictor becomes part of an optimization model. 
Results on synthetic and real-world datasets show that the accuracy of the 
method is comparable to that of alternative approaches that admit a
piecewise linear form of similar complexity. A possible drawback of PARC is 
its computation time, mainly due to solving a sequence of softmax regression 
problems. This makes PARC applicable to datasets whose size, in terms of number 
of samples and features, is such that standard softmax regression is a feasible 
approach.

Other regression and classification methods, such as deep neural networks,
more complex decision trees, and even random forests may achieve better
scores on test data and reduced training time. However, they would return
predictors that are more complicated to optimize over the predictor than the 
proposed piecewise linear models. 

The proposed algorithm can be extended in several ways. For example,    
$\ell_1$-penalties can be introduced in~\eqref{eq:cost_k-y} to promote sparsity 
of $a,b$. The proof of Theorem~\ref{th:convergence} can be easily extended
to cover such a modification. Moreover, basis functions $\phi_i(x)$ can be used 
instead of $x$ directly, such as canonical piecewise linear 
functions~\citep{LU92,CD88,JDD00} to maintain the PWL nature of the 
predictor, with possibly different basis functions chosen for partitioning the 
feature space and for fitting targets.

The proposed algorithm is also extendable to other regression, classification, 
and separation methods than linear ones, as long as we can associate a suitable 
cost function $V^y$/$V^x$. As an example, neural networks with ReLU activation 
functions might be used instead of ridge 
regression for extended flexibility, for which we can define 
$V_y(a^j,b^j,x_k,y_k)$ as the loss computed on the training data of cluster 
\#$j$.

Ongoing research is devoted to alternative methods to obtain the initial
assignment of datapoints to clusters, as this is a crucial step that affects
the quality of the minimum PARC converges to, and to applying the proposed
method to data-driven model predictive control of hybrid dynamical systems.

\begin{table}[h!]\begin{center}
\scalebox{.7}[0.7]{
\begin{tabular}{l|r|r|r|r|r|r|r}
dataset & PARC & PARC  & PARC & PARC
& ridge & NN
& DT
\\
$N_{\rm tot}$, $n_x$, $K^*$ & $K^*$ & $K=3$ & $K=5$ & $K=12$
& $K=1$ & $K^*$
& $10$
\\[.5em]\hline
{\small\textsf{1028\_SWD}}
& 0.466& 0.484& 0.515& \textbf{0.529}& 0.441& 0.423& 0.388\\ 1000, 21, 2 
& (1.2\%)& (1.2\%)& (1.2\%)& (1.2\%)& (1.1\%)& (6.9\%)& (1.8\%)\\\hline
{\small\textsf{1029\_LEV}}
& 0.589& 0.600& 0.612& \textbf{0.623}& 0.577& 0.561& 0.466\\ 1000, 16, 2 
& (1.4\%)& (1.5\%)& (1.5\%)& (1.3\%)& (1.4\%)& (2.3\%)& (1.7\%)\\\hline
{\small\textsf{1030\_ERA}}
& 0.427& 0.427& \textbf{0.427}& 0.427& 0.427& 0.321& 0.347\\ 1000, 51, 2 
& (1.4\%)& (1.4\%)& (1.4\%)& (1.4\%)& (1.4\%)& (10.3\%)& (1.4\%)\\\hline
{\small\textsf{522\_pm10}}
& 0.382& 0.419& 0.515& \textbf{0.768}& 0.246& 0.280& 0.423\\ 500, 29, 2 
& (2.3\%)& (3.1\%)& (3.3\%)& (3.2\%)& (1.8\%)& (7.6\%)& (1.6\%)\\\hline
{\small\textsf{529\_pollen}}
& \textbf{0.796}& 0.794& 0.794& 0.796& 0.793& 0.793& 0.486\\ 3848, 4, 15 
& (0.3\%)& (0.2\%)& (0.2\%)& (0.2\%)& (0.2\%)& (0.3\%)& (0.7\%)\\\hline
{\small\textsf{547\_no2}}
& 0.630& 0.666& 0.706& \textbf{0.855}& 0.559& 0.563& 0.612\\ 500, 29, 2 
& (1.8\%)& (1.8\%)& (1.7\%)& (1.4\%)& (1.7\%)& (3.0\%)& (1.6\%)\\\hline
{\small\textsf{593\_fri\_c1\_1000\_10}}
& 0.766& 0.636& 0.755& \textbf{0.828}& 0.306& 0.689& 0.751\\ 1000, 10, 5 
& (10.3\%)& (7.2\%)& (10.6\%)& (4.1\%)& (1.0\%)& (28.0\%)& (0.9\%)\\\hline
{\small\textsf{595\_fri\_c0\_1000\_10}}
& 0.835& 0.805& 0.836& \textbf{0.893}& 0.722& 0.805& 0.677\\ 1000, 10, 4 
& (2.7\%)& (2.1\%)& (2.4\%)& (1.1\%)& (0.7\%)& (4.8\%)& (1.1\%)\\\hline
{\small\textsf{597\_fri\_c2\_500\_5}}
& 0.934& 0.622& 0.907& \textbf{0.945}& 0.282& 0.930& 0.821\\ 500, 5, 11 
& (1.9\%)& (8.6\%)& (2.4\%)& (2.3\%)& (1.5\%)& (1.2\%)& (0.8\%)\\\hline
{\small\textsf{599\_fri\_c2\_1000\_5}}
& 0.933& 0.698& 0.849& 0.937& 0.312& \textbf{0.942}& 0.791\\ 1000, 5, 10 
& (1.3\%)& (10.7\%)& (7.9\%)& (0.9\%)& (1.0\%)& (0.5\%)& (0.8\%)\\\hline
{\small\textsf{604\_fri\_c4\_500\_10}}
& 0.837& 0.698& 0.829& \textbf{0.891}& 0.297& 0.806& 0.757\\ 500, 10, 7 
& (7.2\%)& (8.2\%)& (7.1\%)& (3.3\%)& (2.1\%)& (20.6\%)& (1.1\%)\\\hline
{\small\textsf{606\_fri\_c2\_1000\_10}}
& 0.766& 0.617& 0.783& \textbf{0.855}& 0.329& 0.488& 0.771\\ 1000, 10, 4 
& (5.4\%)& (10.9\%)& (5.0\%)& (4.4\%)& (1.2\%)& (22.1\%)& (0.9\%)\\\hline
{\small\textsf{608\_fri\_c3\_1000\_10}}
& 0.842& 0.494& 0.854& 0.872& 0.305& \textbf{0.901}& 0.748\\ 1000, 10, 7 
& (4.6\%)& (7.5\%)& (2.8\%)& (3.7\%)& (1.2\%)& (8.6\%)& (1.2\%)\\\hline
{\small\textsf{609\_fri\_c0\_1000\_5}}
& \textbf{0.936}& 0.821& 0.877& 0.934& 0.730& 0.909& 0.676\\ 1000, 5, 15 
& (0.7\%)& (2.8\%)& (2.3\%)& (0.6\%)& (0.8\%)& (1.6\%)& (0.8\%)\\\hline
{\small\textsf{612\_fri\_c1\_1000\_5}}
& 0.909& 0.563& 0.750& 0.898& 0.264& \textbf{0.943}& 0.746\\ 1000, 5, 14 
& (2.6\%)& (24.3\%)& (11.8\%)& (4.2\%)& (0.9\%)& (0.4\%)& (0.7\%)\\\hline
{\small\textsf{617\_fri\_c3\_500\_5}}
& 0.906& 0.820& 0.879& \textbf{0.927}& 0.270& 0.892& 0.780\\ 500, 5, 10 
& (2.8\%)& (6.4\%)& (2.2\%)& (2.1\%)& (1.6\%)& (1.0\%)& (1.2\%)\\\hline
{\small\textsf{623\_fri\_c4\_1000\_10}}
& 0.854& 0.675& 0.852& \textbf{0.887}& 0.300& 0.870& 0.746\\ 1000, 10, 6 
& (5.9\%)& (8.5\%)& (5.0\%)& (2.3\%)& (1.1\%)& (16.2\%)& (1.0\%)\\\hline
{\small\textsf{627\_fri\_c2\_500\_10}}
& 0.711& 0.624& 0.725& \textbf{0.841}& 0.301& 0.455& 0.798\\ 500, 10, 5 
& (7.6\%)& (17.5\%)& (8.3\%)& (4.7\%)& (1.2\%)& (21.1\%)& (1.1\%)\\\hline
{\small\textsf{628\_fri\_c3\_1000\_5}}
& 0.934& 0.550& 0.907& \textbf{0.937}& 0.268& 0.903& 0.738\\ 1000, 5, 7 
& (0.9\%)& (9.7\%)& (1.8\%)& (0.8\%)& (0.9\%)& (7.0\%)& (0.9\%)\\\hline
{\small\textsf{631\_fri\_c1\_500\_5}}
& 0.904& 0.901& 0.777& \textbf{0.916}& 0.294& 0.842& 0.757\\ 500, 5, 9 
& (3.4\%)& (0.8\%)& (10.4\%)& (2.7\%)& (2.0\%)& (18.1\%)& (0.8\%)\\\hline
{\small\textsf{641\_fri\_c1\_500\_10}}
& 0.746& 0.798& 0.768& \textbf{0.823}& 0.288& 0.371& 0.789\\ 500, 10, 3 
& (18.4\%)& (13.5\%)& (6.9\%)& (3.6\%)& (1.6\%)& (21.4\%)& (1.0\%)\\\hline
{\small\textsf{646\_fri\_c3\_500\_10}}
& 0.877& 0.643& 0.886& \textbf{0.894}& 0.357& 0.706& 0.774\\ 500, 10, 5 
& (5.4\%)& (12.3\%)& (2.9\%)& (3.0\%)& (1.9\%)& (23.3\%)& (1.8\%)\\\hline
{\small\textsf{649\_fri\_c0\_500\_5}}
& 0.928& 0.824& 0.893& \textbf{0.936}& 0.738& 0.886& 0.717\\ 500, 5, 10 
& (0.9\%)& (3.3\%)& (1.7\%)& (1.0\%)& (1.1\%)& (2.0\%)& (1.2\%)\\\hline
{\small\textsf{654\_fri\_c0\_500\_10}}
& 0.822& 0.797& 0.825& \textbf{0.890}& 0.700& 0.797& 0.697\\ 500, 10, 5 
& (2.5\%)& (2.3\%)& (2.1\%)& (2.3\%)& (1.3\%)& (4.8\%)& (1.4\%)\\\hline
{\small\textsf{666\_rmftsa\_ladata}}
& 0.660& 0.671& 0.723& \textbf{0.811}& 0.581& 0.525& 0.732\\ 508, 10, 2 
& (4.1\%)& (3.0\%)& (3.2\%)& (2.1\%)& (2.2\%)& (10.3\%)& (2.2\%)\\\hline
{\small\textsf{titanic}}
& 0.278& 0.295& 0.296& 0.279& 0.253& 0.292& \textbf{0.300}\\ 2201, 5, 12 
& (1.1\%)& (1.2\%)& (1.1\%)& (1.1\%)& (1.1\%)& (1.1\%)& (0.8\%)\\\hline
\end{tabular}}
\caption{Real-world datasets for regression: average $R^2$ score (standard deviation) over 20 runs on training data (best result is highlighted in boldface).}
\label{tab:dataset-regression-training}
\end{center}
\end{table}

\begin{table}[h!]\begin{center}
\scalebox{.7}[0.7]{            
\begin{tabular}{l|r|r|r|r|r|r|r}
dataset & PARC & PARC  & PARC & PARC
& ridge & NN
& DT
\\
$N_{\rm tot}$, $n_x$, $K^*$ & $K^*$ & $K=3$ & $K=5$ & $K=12$
& $K=1$ & $K^*$
& $10$
\\[.5em]\hline
{\small\textsf{1028\_SWD}}
& 0.413& 0.403& 0.383& 0.372& \textbf{0.425}& 0.423& 0.334\\ 1000, 21, 2 
& (4.6\%)& (5.0\%)& (4.8\%)& (5.1\%)& (4.6\%)& (6.9\%)& (4.2\%)\\\hline
{\small\textsf{1029\_LEV}}
& 0.536& 0.533& 0.519& 0.510& 0.542& \textbf{0.561}& 0.412\\ 1000, 16, 2 
& (6.2\%)& (6.7\%)& (7.1\%)& (7.6\%)& (6.5\%)& (2.3\%)& (8.3\%)\\\hline
{\small\textsf{1030\_ERA}}
& \textbf{0.339}& 0.339& 0.339& 0.339& 0.339& 0.321& 0.269\\ 1000, 51, 2 
& (6.7\%)& (6.7\%)& (6.7\%)& (6.7\%)& (6.8\%)& (10.3\%)& (6.5\%)\\\hline
{\small\textsf{522\_pm10}}
& 0.095& 0.043& -0.048& -0.896& 0.095& \textbf{0.280}& 0.177\\ 500, 29, 2 
& (12.3\%)& (12.2\%)& (14.4\%)& (61.6\%)& (8.1\%)& (7.6\%)& (11.5\%)\\\hline
{\small\textsf{529\_pollen}}
& 0.793& 0.796& 0.796& 0.793& \textbf{0.796}& 0.793& 0.438\\ 3848, 4, 15 
& (1.0\%)& (1.0\%)& (1.0\%)& (0.9\%)& (1.0\%)& (0.3\%)& (1.7\%)\\\hline
{\small\textsf{547\_no2}}
& 0.478& 0.468& 0.403& -0.189& 0.488& \textbf{0.563}& 0.420\\ 500, 29, 2 
& (9.0\%)& (11.0\%)& (12.2\%)& (32.1\%)& (8.3\%)& (3.0\%)& (7.1\%)\\\hline
{\small\textsf{593\_fri\_c1\_1000\_10}}
& \textbf{0.696}& 0.582& 0.694& 0.693& 0.292& 0.689& 0.671\\ 1000, 10, 5 
& (12.4\%)& (9.1\%)& (12.4\%)& (8.6\%)& (3.8\%)& (28.0\%)& (3.4\%)\\\hline
{\small\textsf{595\_fri\_c0\_1000\_10}}
& 0.804& 0.760& 0.788& \textbf{0.813}& 0.693& 0.805& 0.585\\ 1000, 10, 4 
& (3.8\%)& (4.9\%)& (3.2\%)& (3.8\%)& (3.1\%)& (4.8\%)& (2.9\%)\\\hline
{\small\textsf{597\_fri\_c2\_500\_5}}
& 0.889& 0.570& 0.888& 0.891& 0.274& \textbf{0.930}& 0.701\\ 500, 5, 11 
& (7.0\%)& (11.0\%)& (4.8\%)& (4.6\%)& (6.7\%)& (1.2\%)& (5.3\%)\\\hline
{\small\textsf{599\_fri\_c2\_1000\_5}}
& 0.920& 0.674& 0.828& 0.924& 0.277& \textbf{0.942}& 0.724\\ 1000, 5, 10 
& (1.9\%)& (12.4\%)& (9.8\%)& (1.5\%)& (4.2\%)& (0.5\%)& (2.9\%)\\\hline
{\small\textsf{604\_fri\_c4\_500\_10}}
& 0.579& 0.596& 0.624& 0.433& 0.235& \textbf{0.806}& 0.610\\ 500, 10, 7 
& (23.1\%)& (13.0\%)& (13.6\%)& (42.2\%)& (10.0\%)& (20.6\%)& (6.1\%)\\\hline
{\small\textsf{606\_fri\_c2\_1000\_10}}
& 0.710& 0.575& \textbf{0.725}& 0.700& 0.302& 0.488& 0.712\\ 1000, 10, 4 
& (8.0\%)& (11.0\%)& (7.8\%)& (9.6\%)& (5.4\%)& (22.1\%)& (3.0\%)\\\hline
{\small\textsf{608\_fri\_c3\_1000\_10}}
& 0.766& 0.420& 0.804& 0.729& 0.269& \textbf{0.901}& 0.658\\ 1000, 10, 7 
& (8.5\%)& (9.9\%)& (3.9\%)& (9.5\%)& (5.4\%)& (8.6\%)& (4.9\%)\\\hline
{\small\textsf{609\_fri\_c0\_1000\_5}}
& \textbf{0.918}& 0.811& 0.861& 0.917& 0.725& 0.909& 0.580\\ 1000, 5, 15 
& (1.5\%)& (3.5\%)& (3.6\%)& (1.5\%)& (3.1\%)& (1.6\%)& (3.4\%)\\\hline
{\small\textsf{612\_fri\_c1\_1000\_5}}
& 0.877& 0.524& 0.725& 0.865& 0.256& \textbf{0.943}& 0.690\\ 1000, 5, 14 
& (4.5\%)& (26.0\%)& (12.8\%)& (7.1\%)& (3.4\%)& (0.4\%)& (2.5\%)\\\hline
{\small\textsf{617\_fri\_c3\_500\_5}}
& 0.806& 0.781& 0.814& 0.831& 0.206& \textbf{0.892}& 0.622\\ 500, 5, 10 
& (5.6\%)& (7.8\%)& (7.5\%)& (7.7\%)& (7.2\%)& (1.0\%)& (6.0\%)\\\hline
{\small\textsf{623\_fri\_c4\_1000\_10}}
& 0.775& 0.642& 0.814& 0.766& 0.291& \textbf{0.870}& 0.659\\ 1000, 10, 6 
& (11.6\%)& (10.2\%)& (8.3\%)& (10.7\%)& (4.9\%)& (16.2\%)& (3.9\%)\\\hline
{\small\textsf{627\_fri\_c2\_500\_10}}
& 0.547& 0.521& 0.587& 0.375& 0.252& 0.455& \textbf{0.654}\\ 500, 10, 5 
& (16.5\%)& (23.0\%)& (11.3\%)& (24.0\%)& (6.1\%)& (21.1\%)& (6.0\%)\\\hline
{\small\textsf{628\_fri\_c3\_1000\_5}}
& \textbf{0.928}& 0.554& 0.902& 0.921& 0.278& 0.903& 0.651\\ 1000, 5, 7 
& (2.5\%)& (9.0\%)& (2.2\%)& (2.2\%)& (3.6\%)& (7.0\%)& (2.9\%)\\\hline
{\small\textsf{631\_fri\_c1\_500\_5}}
& 0.857& \textbf{0.883}& 0.735& 0.828& 0.266& 0.842& 0.674\\ 500, 5, 9 
& (5.6\%)& (2.5\%)& (15.3\%)& (7.8\%)& (8.9\%)& (18.1\%)& (6.5\%)\\\hline
{\small\textsf{641\_fri\_c1\_500\_10}}
& 0.658& \textbf{0.730}& 0.659& 0.350& 0.253& 0.371& 0.690\\ 500, 10, 3 
& (27.2\%)& (21.7\%)& (14.7\%)& (19.8\%)& (7.8\%)& (21.4\%)& (4.1\%)\\\hline
{\small\textsf{646\_fri\_c3\_500\_10}}
& 0.767& 0.547& \textbf{0.791}& 0.552& 0.295& 0.706& 0.616\\ 500, 10, 5 
& (10.0\%)& (16.4\%)& (8.3\%)& (15.0\%)& (9.0\%)& (23.3\%)& (6.9\%)\\\hline
{\small\textsf{649\_fri\_c0\_500\_5}}
& 0.881& 0.782& 0.859& 0.874& 0.706& \textbf{0.886}& 0.585\\ 500, 5, 10 
& (3.0\%)& (7.0\%)& (4.0\%)& (3.8\%)& (5.2\%)& (2.0\%)& (6.0\%)\\\hline
{\small\textsf{654\_fri\_c0\_500\_10}}
& 0.708& 0.729& 0.722& 0.604& 0.656& \textbf{0.797}& 0.569\\ 500, 10, 5 
& (5.2\%)& (5.5\%)& (4.3\%)& (15.7\%)& (5.7\%)& (4.8\%)& (6.4\%)\\\hline
{\small\textsf{666\_rmftsa\_ladata}}
& \textbf{0.605}& 0.600& 0.586& 0.424& 0.569& 0.525& 0.436\\ 508, 10, 2 
& (11.0\%)& (7.2\%)& (11.6\%)& (21.5\%)& (7.9\%)& (10.3\%)& (17.7\%)\\\hline
{\small\textsf{titanic}}
& 0.263& 0.280& 0.280& 0.264& 0.248& \textbf{0.292}& 0.273\\ 2201, 5, 12 
& (4.0\%)& (4.3\%)& (4.3\%)& (3.8\%)& (4.4\%)& (1.1\%)& (3.1\%)\\\hline
\end{tabular}}
\caption{Real-world datasets for regression: average $R^2$ score (standard deviation) over 20 runs on test data (best result is highlighted in boldface).}
\label{tab:dataset-regression-test}
\end{center}
\end{table}

\begin{table}[h!]\begin{center}
\scalebox{.9}[0.9]{
\begin{tabular}{l|r|r|r|r|r|r|r}
 & PARC & PARC  & PARC & PARC
& ridge & NN
& DT
\\
dataset & $K^*$ & $K=3$ & $K=5$ & $K=12$
& $K=1$ & $K^*$
& $10$
\\[.5em]\hline
{\small\textsf{1028\_SWD}}
& 0.6666& 1.1096& 1.7139& 3.2273& 0.0006& 0.5088& 0.0005\\\hline
{\small\textsf{1029\_LEV}}
& 0.5621& 1.0421& 1.6887& 3.3071& 0.0005& 0.5423& 0.0007\\\hline
{\small\textsf{1030\_ERA}}
& 0.4840& 1.1526& 2.3644& 3.9153& 0.0007& 0.3873& 0.0021\\\hline
{\small\textsf{522\_pm10}}
& 0.6238& 1.0649& 1.5356& 2.8125& 0.0004& 0.3289& 0.0011\\\hline
{\small\textsf{529\_pollen}}
& 12.2727& 4.3129& 5.6963& 10.3119& 0.0003& 0.3802& 0.0038\\\hline
{\small\textsf{547\_no2}}
& 0.7249& 0.9619& 1.5235& 2.7249& 0.0006& 0.3504& 0.0011\\\hline
{\small\textsf{593\_fri\_c1\_1000\_10}}
& 1.8327& 1.1290& 1.8705& 3.5305& 0.0003& 0.8015& 0.0022\\\hline
{\small\textsf{595\_fri\_c0\_1000\_10}}
& 1.5308& 1.3855& 1.7471& 3.2421& 0.0004& 0.4898& 0.0022\\\hline
{\small\textsf{597\_fri\_c2\_500\_5}}
& 1.3449& 0.4908& 0.7735& 1.3944& 0.0003& 0.4973& 0.0009\\\hline
{\small\textsf{599\_fri\_c2\_1000\_5}}
& 2.5124& 1.1925& 1.5406& 2.8690& 0.0005& 0.6089& 0.0012\\\hline
{\small\textsf{604\_fri\_c4\_500\_10}}
& 1.0416& 0.7536& 0.9042& 1.4882& 0.0005& 0.7006& 0.0010\\\hline
{\small\textsf{606\_fri\_c2\_1000\_10}}
& 1.5651& 1.3613& 1.8140& 3.3661& 0.0004& 0.5473& 0.0021\\\hline
{\small\textsf{608\_fri\_c3\_1000\_10}}
& 2.1771& 1.3441& 1.9553& 3.7203& 0.0004& 0.9239& 0.0020\\\hline
{\small\textsf{609\_fri\_c0\_1000\_5}}
& 3.0561& 1.0232& 1.4174& 2.5315& 0.0003& 0.3431& 0.0012\\\hline
{\small\textsf{612\_fri\_c1\_1000\_5}}
& 3.3496& 1.1189& 1.5416& 2.9122& 0.0003& 0.5835& 0.0012\\\hline
{\small\textsf{617\_fri\_c3\_500\_5}}
& 1.5364& 0.6323& 0.9217& 1.7485& 0.0003& 0.5978& 0.0007\\\hline
{\small\textsf{623\_fri\_c4\_1000\_10}}
& 2.1292& 1.4544& 1.9706& 3.7491& 0.0005& 0.9320& 0.0022\\\hline
{\small\textsf{627\_fri\_c2\_500\_10}}
& 0.7763& 0.7443& 0.8311& 1.2776& 0.0007& 0.4008& 0.0010\\\hline
{\small\textsf{628\_fri\_c3\_1000\_5}}
& 2.1933& 1.2742& 1.7093& 3.3434& 0.0007& 0.7285& 0.0012\\\hline
{\small\textsf{631\_fri\_c1\_500\_5}}
& 1.3440& 0.6531& 0.8145& 1.5589& 0.0004& 0.5135& 0.0006\\\hline
{\small\textsf{641\_fri\_c1\_500\_10}}
& 0.6197& 0.6375& 0.7895& 1.4259& 0.0005& 0.3879& 0.0011\\\hline
{\small\textsf{646\_fri\_c3\_500\_10}}
& 0.8552& 0.6777& 0.8998& 1.4902& 0.0005& 0.5253& 0.0012\\\hline
{\small\textsf{649\_fri\_c0\_500\_5}}
& 1.0869& 0.5163& 0.7222& 1.2470& 0.0005& 0.2539& 0.0000\\\hline
{\small\textsf{654\_fri\_c0\_500\_10}}
& 0.7623& 0.5628& 0.7580& 1.2344& 0.0005& 0.3452& 0.0010\\\hline
{\small\textsf{666\_rmftsa\_ladata}}
& 0.4828& 0.6842& 0.9532& 1.6299& 0.0003& 0.4306& 0.0009\\\hline
{\small\textsf{titanic}}
& 1.3399& 0.8540& 0.9237& 1.3155& 0.0003& 0.3517& 0.0004\\\hline
\end{tabular}}
\caption{Real-world datasets for regression: average CPU time (s)
 over 20 runs on regression data.}
\label{tab:dataset-CPUTIME-regression}
\end{center}
\end{table}

\begin{table}[h!]\begin{center}
\scalebox{.9}[0.9]{
\begin{tabular}{l|r|r|r|r|r|r}
dataset & PARC  & PARC & PARC
& softmax & NN
& DT
\\
$N_{\rm tot}$, $n_x$, $m_1$
& $K=2$ & $K=3$ & $K=5$
& $K=1$ & $5$
& $5$
\\[.5em]\hline
{\small\textsf{car}}
&  0.96&  0.97& \textbf{ 0.98}&  0.95&  0.97&  0.84\\ 1728, 15, 4
& (0.8\%)& (1.0\%)& (1.1\%)& (0.4\%)& (1.0\%)& (0.6\%)\\\hline
{\small\textsf{churn}}
&  0.91&  0.90&  0.90&  0.87&  0.93& \textbf{ 0.94}\\ 5000, 21, 2
& (1.9\%)& (1.3\%)& (1.0\%)& (0.2\%)& (1.3\%)& (0.2\%)\\\hline
{\small\textsf{cmc}}
&  0.58&  0.57& \textbf{ 0.60}&  0.53&  0.59&  0.58\\ 1473, 17, 3
& (1.0\%)& (1.3\%)& (1.4\%)& (0.9\%)& (1.7\%)& (0.7\%)\\\hline
{\small\textsf{contraceptive}}
&  0.59&  0.58& \textbf{ 0.60}&  0.53&  0.59&  0.58\\ 1473, 17, 3
& (1.0\%)& (0.8\%)& (1.0\%)& (0.7\%)& (2.6\%)& (0.7\%)\\\hline
{\small\textsf{credit\_g}}
&  0.79&  0.82& \textbf{ 0.88}&  0.77&  0.85&  0.78\\ 1000, 37, 2
& (0.7\%)& (1.2\%)& (1.3\%)& (0.9\%)& (1.2\%)& (0.9\%)\\\hline
{\small\textsf{flare}}
&  0.84&  0.85& \textbf{ 0.85}&  0.84&  0.84&  0.85\\ 1066, 13, 2
& (0.7\%)& (0.7\%)& (0.9\%)& (0.5\%)& (0.6\%)& (0.7\%)\\\hline
{\small\textsf{GAMETES\_E**0.1H}}
&  0.64&  0.64&  0.70&  0.56& \textbf{ 0.75}&  0.57\\ 1600, 40, 2
& (1.8\%)& (1.3\%)& (1.2\%)& (1.0\%)& (1.1\%)& (1.8\%)\\\hline
{\small\textsf{GAMETES\_E**0.4H}}
&  0.69&  0.72&  0.80&  0.55& \textbf{ 0.84}&  0.54\\ 1600, 38, 2
& (6.3\%)& (4.8\%)& (5.0\%)& (1.0\%)& (0.8\%)& (1.7\%)\\\hline
{\small\textsf{GAMETES\_E**0.2H}}
&  0.62&  0.64& \textbf{ 0.69}&  0.58&  0.68&  0.57\\ 1600, 40, 2
& (1.1\%)& (1.1\%)& (1.1\%)& (0.9\%)& (1.9\%)& (0.7\%)\\\hline
{\small\textsf{GAMETES\_H**\_50}}
&  0.62&  0.64&  0.70&  0.55& \textbf{ 0.77}&  0.57\\ 1600, 39, 2
& (2.0\%)& (2.0\%)& (2.0\%)& (0.8\%)& (1.5\%)& (2.4\%)\\\hline
{\small\textsf{GAMETES\_H**\_75}}
&  0.63&  0.70&  0.73&  0.56& \textbf{ 0.79}&  0.56\\ 1600, 39, 2
& (2.1\%)& (3.5\%)& (3.2\%)& (0.9\%)& (1.3\%)& (2.8\%)\\\hline
{\small\textsf{german}}
&  0.80&  0.83& \textbf{ 0.88}&  0.77&  0.85&  0.78\\ 1000, 37, 2
& (0.8\%)& (1.1\%)& (1.2\%)& (0.8\%)& (1.3\%)& (1.0\%)\\\hline
{\small\textsf{led7}}
&  0.75&  0.75& \textbf{ 0.75}&  0.75&  0.73&  0.69\\ 3200, 7, 10
& (0.6\%)& (0.6\%)& (0.5\%)& (0.6\%)& (1.5\%)& (0.7\%)\\\hline
{\small\textsf{mfeat\_morphological}}
&  0.77&  0.77& \textbf{ 0.77}&  0.76&  0.74&  0.72\\ 2000, 7, 10
& (0.5\%)& (0.5\%)& (0.6\%)& (0.5\%)& (2.1\%)& (0.8\%)\\\hline
{\small\textsf{mofn\_3\_7\_10}}
& \textbf{ 1.00}&  1.00&  1.00&  1.00&  1.00&  0.88\\ 1324, 10, 2
& (0.0\%)& (0.0\%)& (0.0\%)& (0.0\%)& (0.0\%)& (0.5\%)\\\hline
{\small\textsf{parity5+5}}
&  0.55&  0.59& \textbf{ 0.69}&  0.52&  0.61&  0.54\\ 1124, 10, 2
& (2.6\%)& (8.0\%)& (13.8\%)& (1.3\%)& (14.4\%)& (1.4\%)\\\hline
{\small\textsf{segmentation}}
&  0.97&  0.98& \textbf{ 0.98}&  0.97&  0.97&  0.94\\ 2310, 21, 7
& (0.5\%)& (0.5\%)& (0.3\%)& (0.3\%)& (0.5\%)& (0.3\%)\\\hline
{\small\textsf{solar\_flare\_2}}
&  0.79&  0.80& \textbf{ 0.81}&  0.78&  0.78&  0.77\\ 1066, 16, 6
& (0.8\%)& (0.9\%)& (0.8\%)& (0.8\%)& (1.2\%)& (0.9\%)\\\hline
{\small\textsf{wine\_quality\_red}}
&  0.62&  0.64& \textbf{ 0.66}&  0.61&  0.62&  0.62\\ 1599, 11, 6
& (1.0\%)& (1.0\%)& (1.2\%)& (0.5\%)& (1.1\%)& (1.3\%)\\\hline
{\small\textsf{wine\_quality\_white}}
&  0.55&  0.55& \textbf{ 0.57}&  0.54&  0.56&  0.54\\ 4898, 11, 7
& (0.4\%)& (0.6\%)& (0.7\%)& (0.3\%)& (0.4\%)& (0.6\%)\\\hline
{\small\textsf{yeast}}
&  0.61&  0.62& \textbf{ 0.64}&  0.60&  0.60&  0.61\\ 1479, 9, 9
& (0.8\%)& (1.0\%)& (1.1\%)& (0.7\%)& (1.0\%)& (1.0\%)\\\hline
\end{tabular}}
\caption{Real-world datasets for classification: average accuracy score (standard deviation) over 20 runs on training data (best result is highlighted in boldface).}
\label{tab:dataset-classification-training}
\end{center}
\end{table}

\begin{table}[h!]\begin{center}
\scalebox{.9}[0.9]{
\begin{tabular}{l|r|r|r|r|r|r}
dataset & PARC  & PARC & PARC
& softmax & NN
& DT
\\
$N_{\rm tot}$, $n_x$, $m_1$
& $K=2$ & $K=3$ & $K=5$
& $K=1$ & $5$
& $5$
\\[.5em]\hline
{\small\textsf{car}}
&  0.94&  0.94&  0.95&  0.93& \textbf{ 0.95}&  0.81\\ 1728, 15, 4
& (1.7\%)& (1.1\%)& (2.1\%)& (1.2\%)& (2.0\%)& (1.7\%)\\\hline
{\small\textsf{churn}}
&  0.90&  0.89&  0.88&  0.86&  0.92& \textbf{ 0.93}\\ 5000, 21, 2
& (1.6\%)& (1.9\%)& (1.0\%)& (0.8\%)& (1.6\%)& (0.7\%)\\\hline
{\small\textsf{cmc}}
&  0.55&  0.53&  0.52&  0.51&  0.56& \textbf{ 0.56}\\ 1473, 17, 3
& (2.9\%)& (2.9\%)& (2.5\%)& (3.3\%)& (2.6\%)& (2.6\%)\\\hline
{\small\textsf{contraceptive}}
&  0.54&  0.52&  0.52&  0.50&  0.54& \textbf{ 0.56}\\ 1473, 17, 3
& (2.6\%)& (2.1\%)& (2.3\%)& (2.4\%)& (2.9\%)& (2.9\%)\\\hline
{\small\textsf{credit\_g}}
&  0.73&  0.72&  0.69& \textbf{ 0.74}&  0.72&  0.73\\ 1000, 37, 2
& (2.5\%)& (3.3\%)& (2.4\%)& (2.9\%)& (2.8\%)& (2.6\%)\\\hline
{\small\textsf{flare}}
&  0.82&  0.82&  0.82&  0.83& \textbf{ 0.83}&  0.81\\ 1066, 13, 2
& (2.3\%)& (2.4\%)& (2.8\%)& (2.6\%)& (2.2\%)& (2.6\%)\\\hline
{\small\textsf{GAMETES\_E**0.1H}}
&  0.54&  0.51&  0.53&  0.48& \textbf{ 0.62}&  0.50\\ 1600, 40, 2
& (3.6\%)& (2.3\%)& (3.5\%)& (2.7\%)& (2.6\%)& (2.9\%)\\\hline
{\small\textsf{GAMETES\_E**0.4H}}
&  0.62&  0.63&  0.68&  0.47& \textbf{ 0.75}&  0.49\\ 1600, 38, 2
& (8.8\%)& (7.2\%)& (6.5\%)& (2.6\%)& (2.4\%)& (2.7\%)\\\hline
{\small\textsf{GAMETES\_E**0.2H}}
&  0.51&  0.51&  0.51& \textbf{ 0.51}&  0.51&  0.51\\ 1600, 40, 2
& (2.0\%)& (2.2\%)& (2.7\%)& (2.6\%)& (2.3\%)& (2.6\%)\\\hline
{\small\textsf{GAMETES\_H**\_50}}
&  0.52&  0.52&  0.54&  0.49& \textbf{ 0.65}&  0.51\\ 1600, 39, 2
& (3.7\%)& (3.6\%)& (4.4\%)& (2.1\%)& (2.9\%)& (4.9\%)\\\hline
{\small\textsf{GAMETES\_H**\_75}}
&  0.51&  0.61&  0.58&  0.49& \textbf{ 0.68}&  0.51\\ 1600, 39, 2
& (2.9\%)& (4.5\%)& (5.3\%)& (1.8\%)& (3.1\%)& (5.1\%)\\\hline
{\small\textsf{german}}
&  0.73&  0.71&  0.70& \textbf{ 0.74}&  0.73&  0.72\\ 1000, 37, 2
& (3.2\%)& (3.1\%)& (3.3\%)& (3.3\%)& (3.0\%)& (3.0\%)\\\hline
{\small\textsf{led7}}
&  0.73&  0.73&  0.73& \textbf{ 0.73}&  0.72&  0.68\\ 3200, 7, 10
& (2.0\%)& (1.9\%)& (1.9\%)& (1.9\%)& (1.9\%)& (1.7\%)\\\hline
{\small\textsf{mfeat\_morphological}}
& \textbf{ 0.74}&  0.74&  0.74&  0.74&  0.73&  0.69\\ 2000, 7, 10
& (1.9\%)& (2.0\%)& (1.9\%)& (2.1\%)& (3.1\%)& (2.3\%)\\\hline
{\small\textsf{mofn\_3\_7\_10}}
& \textbf{ 1.00}&  1.00&  1.00&  1.00&  1.00&  0.83\\ 1324, 10, 2
& (0.0\%)& (0.0\%)& (0.0\%)& (0.0\%)& (0.0\%)& (1.7\%)\\\hline
{\small\textsf{parity5+5}}
&  0.43&  0.51& \textbf{ 0.60}&  0.44&  0.57&  0.42\\ 1124, 10, 2
& (4.5\%)& (11.9\%)& (18.4\%)& (2.9\%)& (16.1\%)& (3.1\%)\\\hline
{\small\textsf{segmentation}}
&  0.95&  0.95&  0.95& \textbf{ 0.96}&  0.95&  0.93\\ 2310, 21, 7
& (0.9\%)& (1.0\%)& (0.9\%)& (0.9\%)& (0.8\%)& (1.0\%)\\\hline
{\small\textsf{solar\_flare\_2}}
&  0.76&  0.75&  0.74& \textbf{ 0.76}&  0.76&  0.75\\ 1066, 16, 6
& (2.5\%)& (2.5\%)& (2.3\%)& (3.2\%)& (2.7\%)& (3.4\%)\\\hline
{\small\textsf{wine\_quality\_red}}
&  0.58&  0.59&  0.58&  0.59& \textbf{ 0.60}&  0.56\\ 1599, 11, 6
& (2.1\%)& (1.3\%)& (2.9\%)& (2.5\%)& (1.8\%)& (1.9\%)\\\hline
{\small\textsf{wine\_quality\_white}}
&  0.54&  0.54&  0.54&  0.54& \textbf{ 0.54}&  0.52\\ 4898, 11, 7
& (1.4\%)& (1.3\%)& (1.6\%)& (1.2\%)& (1.3\%)& (1.7\%)\\\hline
{\small\textsf{yeast}}
& \textbf{ 0.59}&  0.58&  0.58&  0.59&  0.57&  0.57\\ 1479, 9, 9
& (2.3\%)& (2.5\%)& (2.5\%)& (2.1\%)& (2.6\%)& (3.2\%)\\\hline
\end{tabular}}
\caption{Real-world datasets for classification: average accuracy score (standard deviation) over 20 runs on test data (best result is highlighted in boldface).}
\label{tab:dataset-classification-test}
\end{center}
\end{table}

\begin{table}[h!]\begin{center}
\scalebox{.9}[0.9]{
\begin{tabular}{l|r|r|r|r|r|r}
 & PARC  & PARC & PARC
& softmax & NN
& DT
\\
dataset & $K=2$ & $K=3$ & $K=5$
& $K=1$ & $5$
& $5$
\\[.5em]\hline
{\small\textsf{car}}
& 7.4924& 9.0169& 12.8278& 0.1412& 2.4207& 0.0008\\\hline
{\small\textsf{churn}}
& 13.9831& 22.0390& 36.4420& 0.0743& 2.6273& 0.0159\\\hline
{\small\textsf{cmc}}
& 7.5515& 19.7245& 9.6798& 0.0709& 1.0552& 0.0010\\\hline
{\small\textsf{contraceptive}}
& 6.4513& 20.1851& 9.4557& 0.0690& 1.0272& 0.0011\\\hline
{\small\textsf{credit\_g}}
& 2.3672& 4.7748& 8.7926& 0.0391& 1.3028& 0.0015\\\hline
{\small\textsf{flare}}
& 1.5304& 2.1173& 4.0786& 0.0249& 0.2994& 0.0004\\\hline
{\small\textsf{GAMETES\_E**0.1H}}
& 2.7267& 4.9315& 10.3220& 0.0547& 1.7027& 0.0017\\\hline
{\small\textsf{GAMETES\_E**0.4H}}
& 2.8602& 5.4326& 10.3689& 0.0553& 1.5453& 0.0013\\\hline
{\small\textsf{GAMETES\_E**0.2H}}
& 2.9306& 5.0253& 9.8749& 0.0538& 1.4765& 0.0017\\\hline
{\small\textsf{GAMETES\_H**\_50}}
& 2.8087& 5.2419& 9.8828& 0.0668& 1.7266& 0.0013\\\hline
{\small\textsf{GAMETES\_H**\_75}}
& 2.3300& 5.1630& 10.4145& 0.0482& 1.6610& 0.0016\\\hline
{\small\textsf{german}}
& 2.4934& 4.5721& 8.6540& 0.0431& 1.2316& 0.0015\\\hline
{\small\textsf{led7}}
& 12.2785& 23.7578& 38.3652& 0.2652& 2.7853& 0.0009\\\hline
{\small\textsf{mfeat\_morphological}}
& 41.7433& 38.5157& 30.7540& 0.4099& 3.0499& 0.0025\\\hline
{\small\textsf{mofn\_3\_7\_10}}
& 0.4777& 0.6472& 1.0741& 0.0135& 0.8536& 0.0007\\\hline
{\small\textsf{parity5+5}}
& 1.0691& 2.9022& 5.4212& 0.0067& 0.6364& 0.0005\\\hline
{\small\textsf{segmentation}}
& 44.5708& 31.6451& 28.3626& 0.4591& 2.9740& 0.0091\\\hline
{\small\textsf{solar\_flare\_2}}
& 8.4624& 9.0207& 8.2567& 0.2303& 1.3513& 0.0007\\\hline
{\small\textsf{wine\_quality\_red}}
& 21.8689& 44.5064& 16.4515& 0.1968& 1.3265& 0.0027\\\hline
{\small\textsf{wine\_quality\_white}}
& 46.1583& 65.0352& 104.6381& 0.7758& 3.0010& 0.0072\\\hline
{\small\textsf{yeast}}
& 8.9445& 24.1496& 36.5156& 0.1863& 1.6473& 0.0014\\\hline
\end{tabular}}
\caption{Real-world datasets for classification: average CPU time (s)
 over 20 runs on classification data.}
\label{tab:dataset-CPUTIME-classification}
\end{center}
\end{table}

\end{document}